\documentclass[journal]{IEEEtai_arxiv}
\usepackage{amsmath,amsfonts}
\usepackage{algorithmic}
\usepackage{algorithm}
\usepackage{array}
\usepackage[caption=false,font=footnotesize,labelfont=sf,textfont=sf]{subfig}
\usepackage{textcomp}
\usepackage{stfloats}
\usepackage{url}
\usepackage{verbatim}
\usepackage{graphicx}
\usepackage{xcolor}
\usepackage{cite}
\usepackage{booktabs}
\usepackage{tabularx}
\usepackage{multirow}
\usepackage{balance}

\newcommand{\KD}{\textsf{KD}}

\begin{document}

\title{Informed Machine Learning with Knowledge Landmarks}

\author{Chuyi Dai, Witold Pedrycz,~\IEEEmembership{Life Fellow,~IEEE}, Suping Xu,\\ Ding Liu,~\IEEEmembership{Member,~IEEE}, Lin Shang, and Xianmin Wang%
\thanks{\emph{(Corresponding authors: Suping Xu; Ding Liu.)}}%
\thanks{Chuyi Dai is with the Department of Electrical and Computer Engineering, University of Alberta, Edmonton, T6G 1H9, Canada.}%
\thanks{Witold Pedrycz is with the Department of Electrical and Computer Engineering, University of Alberta, Edmonton, AB T6G 2R3, Canada, Systems Research Institute, Polish Academy of Sciences, 00-901 Warsaw, Poland, and Research Center of Performance and Productivity Analysis, Istinye University, Istanbul, Türkiye.}%
\thanks{Suping Xu is with the School of Management, Nanjing University of Posts and Telecommunications, Nanjing 210003, China, also with the Department of Electrical and Computer Engineering, University of Alberta, Edmonton, AB T6G 2R3, Canada, and also with the State Key Laboratory of Novel Software Technology, Nanjing University, Nanjing 210023, China.}%
\thanks{Ding Liu is with the School of Mechano-Electronic Engineering, Xidian University, Xi’an 710071, China.}%
\thanks{Lin Shang is with the School of Computer Science, Nanjing University, Nanjing 210023, China, and also with the State Key Laboratory of Novel Software Technology, Nanjing University, Nanjing 210023, China.}%
\thanks{Xianmin Wang is with the Hubei Subsurface Multi-Scale Imaging Key Laboratory, Institute of Geophysics and Geomatics, China University of Geosciences, Wuhan 430074, China.}%
}

\markboth{IEEE Transactions on Artificial Intelligence}
{Informed Machine Learning with Knowledge Landmarks}

\IEEEaftertitletext{%
  {\normalfont\footnotesize\noindent
  \textcopyright~2026 IEEE. Personal use of this material is permitted.
  Permission from IEEE must be obtained for all other uses, in any current
  or future media, including reprinting/republishing this material for
  advertising or promotional purposes, creating new collective works,
  for resale or redistribution to servers or lists, or reuse of any
  copyrighted component of this work in other works.\par}%
  \medskip
}

\maketitle

\begin{abstract}
Informed Machine Learning (ML) aims to improve prediction by using both data and prior knowledge. In scientific and engineering applications, accurate measurements may occupy only a small operating region of the input space, whereas available knowledge is global, coarse, and not expressed as equations. Existing physics-informed methods embed governing equations, residuals, simulators, operators, or mathematically specified constraints. However, they do not directly address learning from localized numeric data together with non-equational knowledge. We develop a Knowledge--Data (\KD) ML framework for this setting. The \KD-model represents coarse global knowledge as input--output information granules, termed knowledge landmarks. The method uses a landmark-guided objective that converts these landmarks into differentiable compatibility terms and balances them against local data fitting through a trade-off parameter $\lambda$. Experiments on two physics-based benchmarks show that the \KD-model predicts more accurately across the full domain than a model trained on the local data alone. For most observation windows, the gain holds over a broad range of $\lambda$. The advantage persists when the sample size or label noise changes, with $\lambda$ re-selected without full-domain labels. These results show that approximate expert knowledge can guide training when dense labels or explicit equations are unavailable.
\end{abstract}

\begin{IEEEImpStatement}
In many scientific and engineering problems, measurements are costly or cover only a small operating region. This work shows how broad but imprecise knowledge can still guide learning. The \KD-model converts such knowledge into landmarks, which describe how inputs and outputs tend to relate, and uses them to predict beyond the observed region without requiring equations or dense measurements. In two controlled physics-based benchmarks, the \KD-model produced more accurate predictions across the full operating range. Where measurements are costly and coarse knowledge is available, the approach may help reduce data-collection and simulation effort.
\end{IEEEImpStatement}


\section{Introduction}
\label{sec:intro}

\IEEEPARstart{M}{achine} learning provides flexible function approximation for scientific and engineering problems~\cite{10.1145/3514228}. Obtaining training data can nevertheless be expensive in some applications~\cite{10090238}. Here, accurate numeric observations cover only a \emph{local} operating (input) region, whereas predictions are needed over a broader domain. The accuracy of predictions outside the observed region is therefore uncertain when a model is trained only on these local observations.

At the same time, useful physical knowledge may come from domain expertise or evidence accumulated across related scenarios. For example, prior physical knowledge may indicate that specific local operating conditions consistently correspond to outputs within a predictable range, even as the exact governing parameters vary across scenarios. Knowledge of this kind is typically coarse-grained and imprecise; it captures general behavioral tendencies. Such knowledge must be encoded into a form that can guide model training alongside localized observations.

Physics-informed machine learning (PIML)  incorporates domain knowledge into learning systems~\cite{karniadakis2021physics,10.1145/3514228,9429985}. Physics-informed neural networks (PINNs), for instance, encode governing PDEs as soft constraints by penalizing differential residuals via automatic differentiation~\cite{raissi2019physics}. Such residual-based formulations require explicit equations for point-wise evaluation. In many engineering tasks, however, available knowledge consists of qualitative or aggregate descriptions, such as operational limits or behavioral envelopes~\cite{10.1145/3514228,10.1115/1.4064449,9429985}. Here, we study prediction from localized observations and such coarse knowledge, with no governing equations supplied. This knowledge must guide learning beyond the observed region.

Granular Computing (GrC) provides a foundation for addressing this gap~\cite{zadeh1997toward,6479257,4385281}. It represents uncertain input--output relationships through \emph{information granules} that reflect variability in system behavior. Within GrC, fuzzy sets connect qualitative system-level descriptions with numeric optimization. Smooth membership functions~\cite{pedrycz2018granular} express degrees of compatibility, so granular guidance can support gradient-based training.
The principle of justifiable granularity (PJG)~\cite{pedrycz2013building,pedrycz2025granular} guides the construction of these granules by balancing empirical coverage and semantic specificity.

This paper proposes a Knowledge--Data (\KD) machine learning framework that combines localized numeric observations with coarse physical knowledge. Its central construct is a collection of \emph{knowledge landmarks} that summarize system behavior across related scenarios, keeping regional input--output associations and omitting point-wise numeric labels. The granularity of each landmark reflects the variability of its associated response regime. Unlike the residual-based methods described above, the \KD-model uses these granular associations as training guidance in place of governing equations. A single objective balances local data fitting with landmark compatibility to guide predictions beyond the observed region. The main contributions are as follows:

\begin{itemize}
    \item We introduce knowledge landmarks that represent coarse physical knowledge as input--output information granules, with a justifiable-granularity rule for selecting input widths.

    \item We develop a differentiable \KD\ learning objective that combines local data fitting with landmark compatibility to guide predictions beyond the observed region.

    \item Experiments on two physics-based benchmarks demonstrate improved full-domain prediction over a data-only baseline. Further analyses examine sensitivity to observation noise and corrupted landmarks.
\end{itemize}

The paper is organized as follows. Section~\ref{sec:problemformulation} formalizes the problem. Section~\ref{sec:literaturereview} reviews the related work. Section~\ref{sec:preliminaries} explains the construction of knowledge landmarks. Section~\ref{sec:lossfunction} presents the augmented loss function and the design process. Section~\ref{sec:experiments} reports experimental results on two physics-based benchmarks. Section~\ref{sec:conclusion} concludes the paper and discusses several future research directions.

\section{Problem Formulation}
\label{sec:problemformulation}

\begin{figure}[!t]
\centering

\subfloat[]{%
  \includegraphics[height=0.42\linewidth]{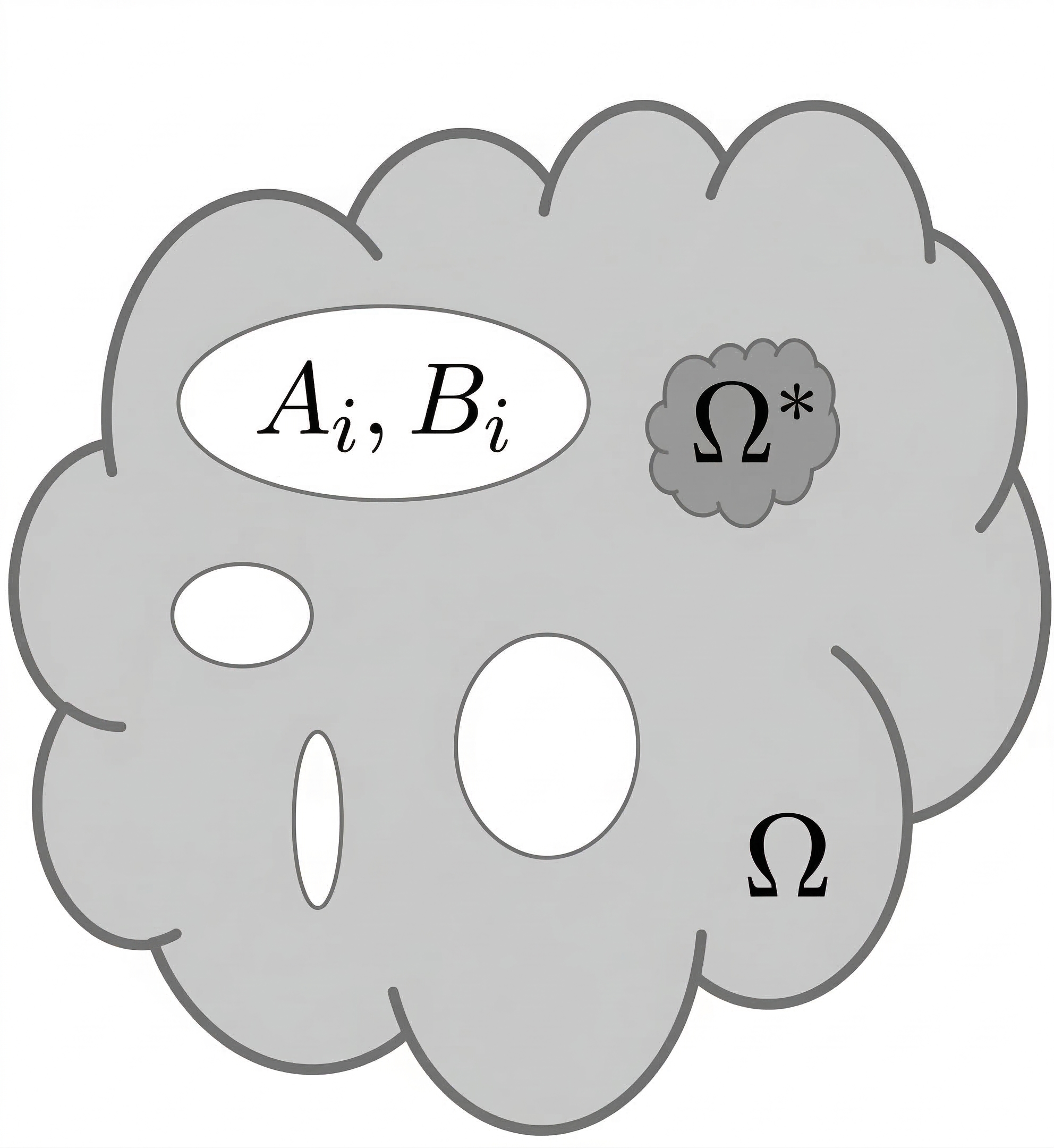}%
  \label{fig:knowledge_landmarks_a}}%
\hspace{0.05\linewidth}%
\subfloat[]{%
  \includegraphics[height=0.42\linewidth]{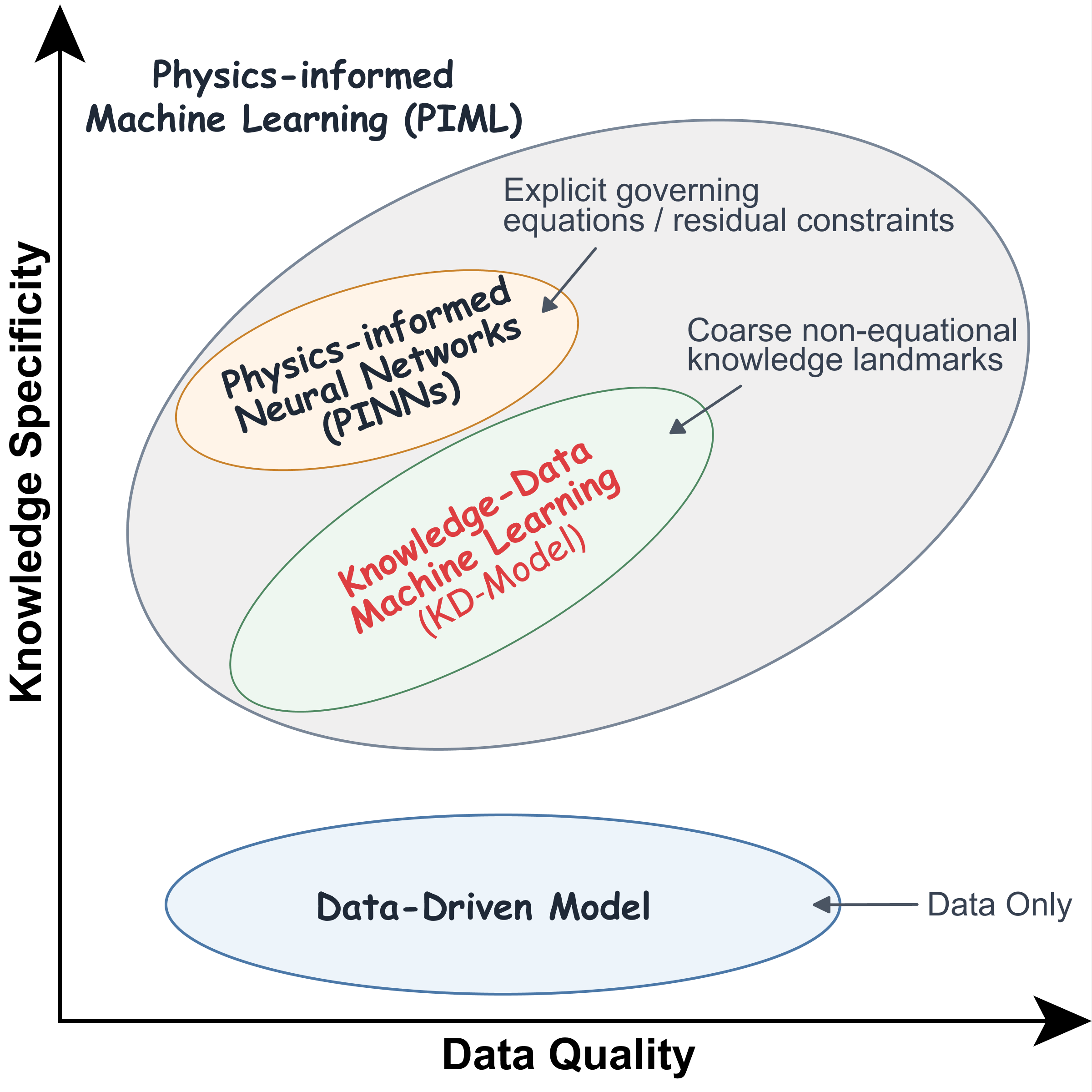}%
  \label{fig:knowledge_landmarks_b}}
\caption{Problem setting and conceptual positioning of the \KD-model. (a) Input space $\Omega$ with knowledge landmarks $(A_i,B_i)$ and localized numeric observations confined to $\Omega^\ast$. (b) Conceptual positioning of the \KD-model relative to PINNs and data-driven models along the qualitative dimensions of knowledge specificity and data quality.}
\label{fig:knowledge_landmarks}
\end{figure}

We consider a machine learning problem in which localized numeric observations are available from an underlying physical system over a limited region of the input domain. We denote the input--output mechanism of this system by $f$:
\begin{equation}
y=f(\boldsymbol{x};\boldsymbol{w}),
\end{equation}
where $\boldsymbol{x}\in\Omega$ denotes the input and $\boldsymbol{w}$ is a vector of parameters whose precise (numeric) values are unknown. The analytical form of $f$ is not used directly in training the \KD-model. In the controlled benchmarks, $f$ is used to generate local observations, prior evidence, and reference outputs for validation and testing. The available numeric data governed by this relation are confined to a local region of the input space, denoted by $\Omega^\ast\subset\Omega$. These localized observations constitute the training dataset $D^\ast$, consisting of $N_1$ input--output pairs:
\begin{equation}
\begin{aligned}
D^\ast &= \left\{\left(\boldsymbol{x}_k,{\mathrm{target}}_k\right)\right\},\\
\boldsymbol{x}_k &\in \Omega^\ast\subset\mathbb{R}^n,\quad k=1,2,\ldots,N_1.
\end{aligned}
\end{equation}
Across different scenarios, locations, and time periods, the data are assumed to reflect the same physical mechanism with varying parameters. We describe this variability through a granular generalization of the physical model,
\begin{equation}
y=f\left(\boldsymbol{x};\boldsymbol{W}\right),
\end{equation}
where the granular vector $\boldsymbol{W}=(w_1,\ldots,w_p)$ expresses admissible parameter variability across related scenarios. This description leads to the knowledge landmarks \textsf{K}, which describe input--output relationships:
\begin{equation}
\mathsf{K}=\left\{(A_i,\ B_i)\right\},\ i=1,2,\ldots,c
\end{equation}
where $A_i$ is an information granule in the input space $\Omega$ and $B_i$ is its corresponding granule located in the output space. The granularity of these pairs reflects parameter variability across scenarios. The landmarks $\mathsf{K}$ are coarse-grained input--output information granules extracted from multi-scenario evidence, not point-wise full-domain labels. They provide coarse guidance across $\Omega$ through graded associations between input and output granules. In contrast, $D^\ast$ provides precise point-wise supervision only within $\Omega^\ast$, as illustrated in Fig.~\ref{fig:knowledge_landmarks}(a). Fig.~\ref{fig:knowledge_landmarks}(b) places the \KD-model within physics-informed ML, using data quality and knowledge specificity as qualitative dimensions.

The design objective is to construct an ML model $M(\boldsymbol{x};\boldsymbol{a})$ with optimized parameters $\boldsymbol{a}$ using both the local data $D^\ast$ and the global knowledge landmarks \textsf{K}. This is achieved by minimizing a composite loss function:
\begin{equation}
L(\boldsymbol{a};\lambda)=\lambda L_1+(1-\lambda)L_2,\ \ \lambda\in[0,1],
\label{eq:lossfunction}
\end{equation}
where the data term $L_1$ measures the fit to the local observations $D^\ast$. The knowledge term $L_2$ is a soft penalty that encourages predictions to remain compatible with the knowledge landmarks. The hyperparameter $\lambda\in[0,1]$ controls the relative weight assigned to local data fitting and granular knowledge regularization. For a fixed value of $\lambda$, the model parameters are obtained through the minimization of the loss function $L$,
\begin{equation}
\boldsymbol{a}_{\lambda}
=
\operatorname{arg\,min}_{\boldsymbol{a}} L(\boldsymbol{a};\lambda),
\qquad \lambda\in[0,1].
\end{equation}
The value of $\lambda$ is selected by the validation process discussed in Section~\ref{sec:lossfunction}. The resulting model is parameterized by $\boldsymbol{a}_{\lambda_{\mathrm{opt}}}$.

\section{Literature Review}
\label{sec:literaturereview}

\subsection{Physics-Informed and Knowledge-Guided Learning}

Physics-informed machine learning aims to improve data efficiency, predictive capability, and the physical plausibility of ML models by embedding domain knowledge into the learning process \cite{karniadakis2021physics,10.1145/3514228,9429985,meng2025piml_survey}. This knowledge may take the form of governing equations, simulation outputs, algebraic relations, logical rules, or expert constraints. These forms can be integrated at different stages of the learning pipeline \cite{10.1145/3514228,9429985,meng2025piml_survey}. Physics-Informed Neural Networks are an established equation-based framework. PINNs incorporate physical laws by penalizing PDE residuals, together with boundary and initial conditions, through automatic differentiation \cite{raissi2019physics}. This formulation has been effective in many forward and inverse problems, especially when the governing equations are known and can be evaluated over the domain. Constructing these residuals requires the form of the differential operator to be specified. PINN training also balances several competing objectives, including data fitting, residual, and boundary terms. This has motivated work on adaptive loss balancing and stabilization \cite{wang2021understanding,bischof2025multiobjective,krishnapriyan2021pinn_failures}, alongside general weighting methods such as GradNorm and SoftAdapt \cite{chen2018gradnorm,heydari2019softadapt}.

Beyond PINNs, a broader class of physics- and knowledge-guided methods incorporates prior knowledge through simulator outputs, probabilistic structures, algebraic constraints, or hybrid data--model couplings \cite{SUBRAMANIAN2023108899,10.1115/1.4064449}. Equation-discovery methods identify governing relationships directly from sparse or noisy observations \cite{BOTH2021109985,chen2021physics}. Operator-learning methods such as DeepONet and neural operators learn mappings between function spaces \cite{lu2021deeponet_nmi,kovachki2023neural_operator_jmlr,azizzadenesheli2024neuraloperators_nrp}. Differentiable constraint-based learning and semantic or logic-based approaches translate symbolic prior knowledge into constraints for gradient-based optimization \cite{roychowdhury2021constraintbased,vankrieken2020fuzzy_implications,xu2018semanticloss,diligenti2017semantic,badreddine2022ltn}. These developments show that useful physical guidance need not be expressed as a classical PDE residual.

\subsection{Learning from Coarse and Imprecise Supervision}

Coarse and imprecise supervision provides learning signals when precise labels are unavailable. Learning from label proportions infers instance-level predictors from aggregate class proportions, while multiple-instance learning uses labels assigned to bags rather than individual instances \cite{quadrianto2009llp,yu2013propsvm,dietterich1997mil}. Weak supervision considers labels that may be incomplete, inexact, or inaccurate \cite{zhou2018weak}. Snorkel combines weak supervision sources through labeling functions \cite{ratner2020snorkel}. Fuzzy regression can represent uncertainty through fuzzy model parameters \cite{tanaka1982fuzzyregression}, while imprecise or interval-valued regression uses set-valued observations \cite{couso2014setvalued,tretiak2023imprecise_regression}.

These studies reduce the need for precise point-wise labels. In these methods, the imprecise information is attached to observed instances, groups of instances, or the fitted model, whereas the constraint-based methods above treat the knowledge as formulas over given predicates. A knowledge landmark can be written as a fuzzy rule, ``if $\boldsymbol{x}$ is $A_i$, then $y$ is $B_i$,'' but fuzzy-logic losses~\cite{vankrieken2020fuzzy_implications,badreddine2022ltn} take the rules and their membership functions as given. The present setting adds three requirements. The granules and their granularity must be built from multi-scenario evidence. The granularity of each landmark must determine how strongly it constrains the prediction. When several landmarks are active at the same unlabeled input, their guidance must be combined into one prediction beyond the observed region.

\subsection{Granular Computing and Justifiable Granularity}

Granular Computing (GrC) formalizes the idea of abstraction by representing imprecise concepts and heterogeneous evidence in the form of information granules \cite{zadeh1997toward,6479257,pedrycz2013granular}. Within GrC, fuzzy sets connect symbolic descriptions with numerical computation through membership degrees \cite{ZADEH1965338,4385281,pedrycz2018granular,11478341}, and the principle of justifiable granularity constructs granules by balancing data support (coverage) against semantic precision (specificity) \cite{pedrycz2013building}.

In ML, GrC has been deployed to extract robust data representations for tasks such as time-series forecasting \cite{9697907} and complex pattern classification \cite{BEHZADIDOOST2024119746}. In these applications, the granules summarize the training data rather than supply separate knowledge that constrains a predictor where data are absent. Many existing implementations treat granular constructs as isolated preprocessing steps \cite{chen2019granular,FU2020105500,niu2022dynamic}. Recent theoretical perspectives position GrC as a foundation for knowledge--data learning and suggest incorporating physical guidance via granular regularization \cite{pedrycz2025granular}. Making this idea operational requires a gradient-based procedure that builds the granules from evidence, scores numeric predictions against them, and combines the data and knowledge terms of the loss. The \KD-model provides such a procedure.

\section{Construction of Knowledge Landmarks}
\label{sec:preliminaries}

We construct knowledge landmarks from prior input--output evidence $D=\{(\boldsymbol{x}_k,y_k)\}_{k=1}^{N}$ across related scenarios. We first form output contexts, then identify the input regions associated with each context. Each input--output granule pair becomes a landmark.

\subsection{The Principle of Justifiable Granularity}

The principle of justifiable granularity constructs an information granule by balancing its data support against the precision (specificity) of its description. Consider one-dimensional data $\{x_1, x_2, \ldots, x_N\}$ and an information granule $A$ with membership function parameterized by a vector $\boldsymbol{p}$. The coverage of $A$ is the normalized sum of the membership degrees it assigns to the data,
\begin{equation}
\operatorname{cov}(A) = \frac{1}{N}\sum_{k=1}^{N} A(x_k).
\label{eq:cov_def}
\end{equation}
For an interval information granule $A=[a,b]$, so that $\boldsymbol{p}=[a,b]$ and $A$ is the corresponding indicator function, Eq.~\eqref{eq:cov_def} reduces to the proportion of samples falling inside the interval.
Specificity decreases as a granule becomes larger. Its form depends on the chosen size measure and calibration. We express it through a decreasing function $\phi$,
\begin{equation}
\operatorname{sp}(A) = \phi\bigl(\operatorname{size}(A)\bigr).
\end{equation}
For an interval granule, $\phi$ can be a linearly decreasing function of the interval length, namely,
\begin{equation}
\operatorname{sp}(A) = 1 - \frac{b-a}{\operatorname{range}},
\label{eq:sp_interval}
\end{equation}
where \textit{range} is a calibration factor set from the data magnitude.
For a fuzzy set, the lengths of its $\alpha$-cuts measure its size. We integrate the $\alpha$-cuts of $A$,
\begin{equation}
\operatorname{sp}(A) = \int_{0}^{1} \operatorname{sp}(A_{\alpha})\, d\alpha.
\label{eq:sp_alpha}
\end{equation}
We balance coverage and specificity by selecting the parameters that maximize the product of coverage and specificity,
\begin{equation}
\boldsymbol{p}_{\mathrm{opt}} = \operatorname{arg\,max}_{\boldsymbol{p}} \, \operatorname{cov}(A)\operatorname{sp}(A).
\end{equation}

\subsection{Construction of Output Contexts}

We first represent the outputs in $D$ with $C$ overlapping contexts $\{B_i\}_{i=1}^{C}$. Each context covers a range of outputs and conditions the input clustering that follows. We use fuzzy Gaussian membership functions,
\begin{equation}
B_i(y) = \exp\left(-\frac{(y-c_i)^2}{\sigma_i^2}\right),
\label{eq:context_gaussian}
\end{equation}
where $c_i$ and $\sigma_i$ denote the center and spread of granule $B_i$.

We use a density-adaptive strategy to balance coverage and specificity when setting the granule parameters. We place the centers $\{c_i\}_{i=1}^{C}$ uniformly between the $2.5$th and $97.5$th percentiles of the outputs in $D$, denoted by $y_{\mathrm{low}}$ and $y_{\mathrm{high}}$. These bounds reduce the influence of extreme values on center placement.
Let $\mathcal{Y}$ contain the $N_{\mathcal{Y}}$ reference outputs used for width estimation. For each center $c_i$, the distance to its $\kappa$th nearest neighbor in $\mathcal{Y}$ is $\delta_i = \bigl|c_i - y_{\mathrm{NN}}^{(\kappa)}\bigr|$, and we set $\sigma_i=\delta_i$. The rank $\kappa=\lceil\rho N_{\mathcal{Y}}\rceil$ is set by a neighborhood fraction $\rho\in(0,1)$, and $\lceil\cdot\rceil$ denotes the ceiling function. Section~\ref{sec:experiments} specifies the reference samples and neighbor ranks used in each benchmark.
The $\alpha$-cut of $B_i$ has length $2\sigma_i\sqrt{\ln(1/\alpha)}$, so Eq.~\eqref{eq:sp_alpha} with the linear score of Eq.~\eqref{eq:sp_interval} gives
\begin{equation}
\operatorname{sp}(B_i)
=
\max\Bigl\{0,\ 1-\frac{\sqrt{\pi}\,\sigma_i}{y_{\mathrm{high}}-y_{\mathrm{low}}}\Bigr\}.
\label{eq:context_specificity}
\end{equation}
The maximum keeps the specificity nonnegative. This value is fixed during training and decreases as the context becomes wider.

\subsection{Construction of Context-Conditioned Input Granules}
\label{subsec:input_granules}

A single output context may correspond to several input regions. For each $B_i$, conditional fuzzy C-means (FCM) finds $K_i$ input prototypes $\boldsymbol{v}_{ij}$ by minimizing the objective function
\begin{equation}
J_i = \sum_{k=1}^{N}\sum_{j=1}^{K_i} u_{ijk}^{\,m}\,\lVert \boldsymbol{x}_k - \boldsymbol{v}_{ij} \rVert^2,
\end{equation}
over memberships $u_{ijk}$ satisfying
\begin{equation}
\sum_{j=1}^{K_i} u_{ijk} = B_i(y_k), \qquad k=1,\ldots,N,
\end{equation}
where $m>1$ is the fuzzification parameter and $U=[u_{ijk}]$ is the partition over contexts, clusters, and samples. The constraint assigns each sample a total membership $B_i(y_k)$ across the input clusters of context $i$. Thus, samples whose outputs better match $B_i$ contribute more membership to its input clusters. The numbers $C$ and $K_i$ are hyperparameters that set the resolution of the landmark set.
The objective function is minimized with respect to the partition and the prototypes. These two sets of variables are updated alternately:
\begin{equation}
u_{ijk} =
\frac{B_i(y_k)}
{\displaystyle \sum_{\ell=1}^{K_i}
\left(
\frac{\lVert \boldsymbol{x}_k - \boldsymbol{v}_{ij} \rVert}
{\lVert \boldsymbol{x}_k - \boldsymbol{v}_{i\ell} \rVert}
\right)^{\frac{2}{m-1}}},
\label{eq:cfcm_u}
\end{equation}
\begin{equation}
\boldsymbol{v}_{ij} =
\frac{\displaystyle \sum_{k=1}^{N} u_{ijk}^{\,m}\boldsymbol{x}_k}
{\displaystyle \sum_{k=1}^{N} u_{ijk}^{\,m}}.
\label{eq:cfcm_v}
\end{equation}

To turn each prototype into an input granule, we assign a width along each coordinate. The supporting evidence has normalized weights
\begin{equation}
w_{ijk} = \frac{u_{ijk}^{\,m}}{\displaystyle \sum_{r=1}^{N} u_{ijr}^{\,m}}, 
\qquad k=1,\ldots,N.
\end{equation}

For width selection, both samples and prototypes are expressed in coordinates normalized to $[0,1]$. Along coordinate $d=1,\ldots,n$, a Gaussian granule centered at $v_{ij}^{(d)}$ has width $\sigma_{ij}^{(d)}\in(0,1]$ and membership
\begin{equation}
\mu_{ij}^{(d)}(x) =
\exp\left(
-\frac{(x-v_{ij}^{(d)})^2}{(\sigma_{ij}^{(d)})^2}
\right).
\end{equation}

Given the normalized support weights $\{w_{ijk}\}_{k=1}^{N}$ induced by conditional FCM, we define the coverage of the candidate one-dimensional granule as the empirical expectation of its membership function with respect to this discrete weighted evidence,
\begin{equation}
\operatorname{cov}_{ij}^{(d)}(\sigma)
=
\sum_{k=1}^{N} w_{ijk}\,\mu_{ij}^{(d)}(x_k^{(d)}).
\end{equation}
We choose a specificity measure based on the normalized width $\sigma\in(0,1]$, $\operatorname{sp}(\sigma) = 1-\sigma$. Under the principle of justifiable granularity, the width is selected by
\begin{equation}
\sigma_{ij}^{(d)}
=
\operatorname{arg\,max}_{\sigma \in (0,1]}
\operatorname{cov}_{ij}^{(d)}(\sigma)\,\operatorname{sp}(\sigma).
\label{eq:width_rule}
\end{equation}

We combine the memberships through the product t-norm,
\begin{equation}
A_{ij}(\boldsymbol{x}) = \prod_{d=1}^{n} \mu_{ij}^{(d)}(x^{(d)}).
\label{eq:input_granule_tnorm}
\end{equation}

The function $A_{ij}(\boldsymbol{x})$ gives the activation at any query input, whereas $u_{ijk}$ describes the membership of an evidence sample during construction. Pairing $A_{ij}$ with its inducing context gives
\begin{equation}
\mathsf{K}
=
\left\{(A_{ij},B_i)\mid i=1,\ldots,C,\ j=1,\ldots,K_i\right\}.
\end{equation}
This yields a total of $c=\sum_{i=1}^{C}K_i$ landmarks. In the loss function, these pairs are flattened and indexed by a single subscript as $\mathsf{K}=\{(A_i, B_i)\}_{i=1}^c$.

\subsection{Matching Information Granules and Numeric Results}
\label{sec:matchinginformation}

\begin{figure}[!t]
\centering
\includegraphics[width=0.70\linewidth]{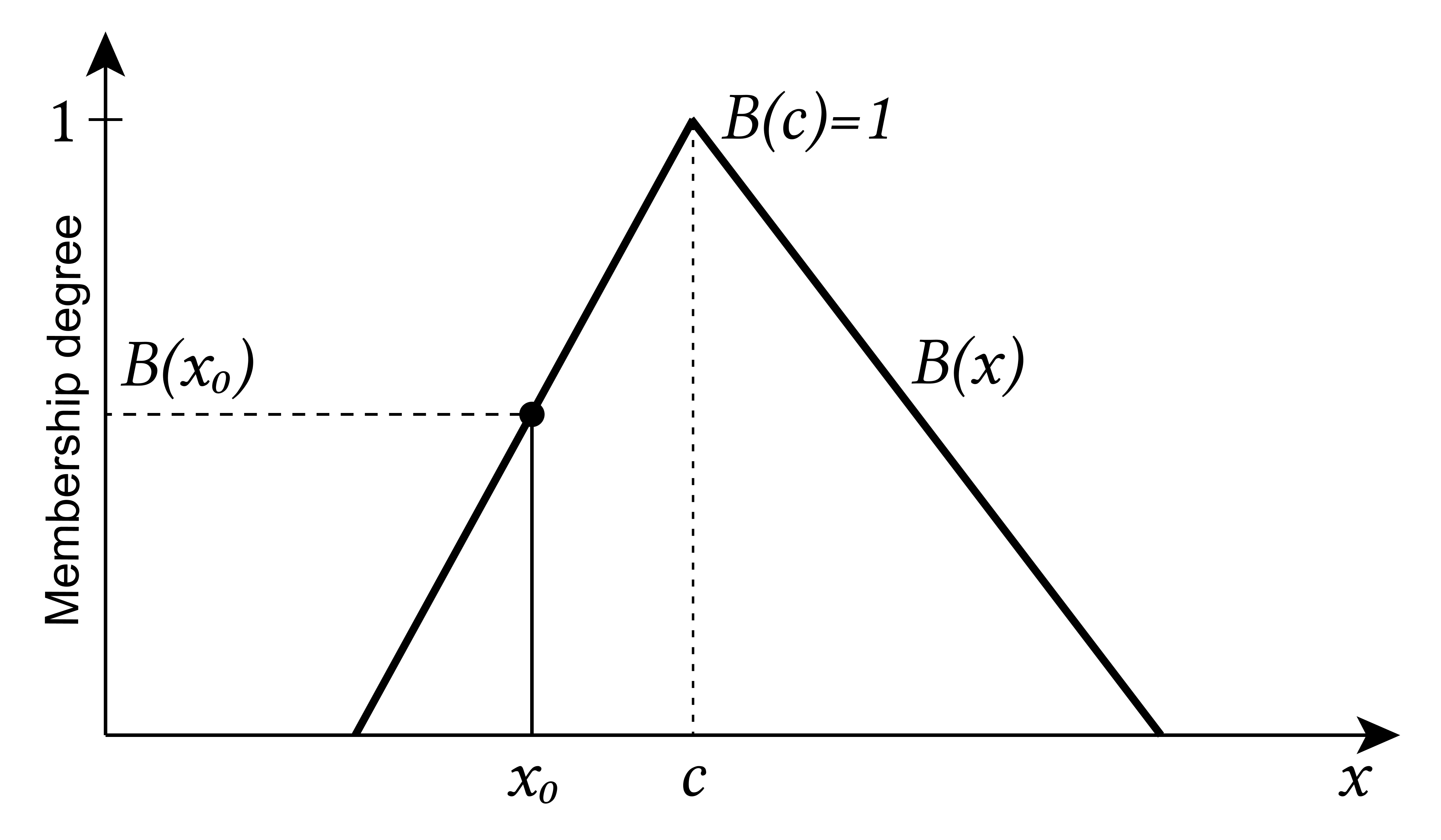}
\caption{Matching of an information granule $B$ and a numeric entity $x_0$.}
\label{fig:matching}
\end{figure}

To use a landmark in training, we evaluate how well a numeric prediction matches its output context. In Fig.~\ref{fig:matching}, $B$ is the granule and $x_0$ denotes a scalar output, representing the prediction $M(\boldsymbol{x};\boldsymbol{a})$. Since the coverage of a point $x_0$ is its membership degree, $\operatorname{cov}(x_0,B)=B(x_0)$, the matching score is defined as $\operatorname{cov}(x_0,B)\operatorname{sp}(B)=B(x_0)\operatorname{sp}(B)$. The specificity factor distinguishes granules that assign $x_0$ the same positive membership: if $B(x_0)=B'(x_0)>0$ and $\operatorname{sp}(B)>\operatorname{sp}(B')$, the match with $B$ is scored higher because it gives a more specific output description. For each context $B_i$, the score is maximized at $x_0=c_i$, reaching $\operatorname{sp}(B_i)$. Section~\ref{sec:lossfunction} applies this score to $M(\boldsymbol{x};\boldsymbol{a})$ and weights its contribution by the input activation $A_i(\boldsymbol{x})$.

\section{Augmented Loss Function, Minimization, and Validation Protocol}
\label{sec:lossfunction}

Following the \KD\ formulation, we develop a training objective that combines localized numeric observations with global granular knowledge. The resulting objective balances empirical data fitting on $D^\ast$ against full-domain landmark compatibility through a trade-off parameter $\lambda$. We first formulate the composite loss and analyze its optimization properties. We then describe the validation protocol used to select $\lambda$ and summarize the complete procedure in Algorithm~\ref{alg:kdml}.

\subsection{Loss Function, Its Interpretation, and Minimization}

The augmented training objective combines empirical data fitting with granular knowledge regularization:
\begin{equation}
L(\boldsymbol{a};\lambda)
=
\lambda L_{\mathrm{data}}(\boldsymbol{a})
+
(1-\lambda)L_{\mathrm{knowledge}}(\boldsymbol{a}),
\qquad \lambda \in [0,1].
\label{eq:augmented_loss}
\end{equation}

The first component of the loss function is a normalized mean-squared error computed on the local data $D^\ast$:
\begin{equation}
L_{\mathrm{data}}(\boldsymbol{a})
=
\frac{1}{N_1 y_{\mathrm{span}}^2}
\sum_{k=1}^{N_1}
\left(M(\boldsymbol{x}_k;\boldsymbol{a}) - \mathrm{target}_k\right)^2,
\label{eq:data_loss}
\end{equation}
where $y_{\mathrm{span}}$ denotes a normalization factor equal to the range of the output variable $\mathrm{target}_k$, $k=1,2,\ldots,N_1$.
The second component, $L_{\mathrm{knowledge}}$, is a soft granular regularizer. Evaluating the compatibility between predictions and output contexts yields
\begin{equation}
L_{\mathrm{knowledge}}(\boldsymbol{a})
=
\frac{1}{N_2}
\sum_{q=1}^{N_2}
\sum_{i=1}^{c}
A_i(\boldsymbol{x}_q)
\left(1 - V_i(\boldsymbol{x}_q;\boldsymbol{a})\right)^2.
\label{eq:knowledge_loss}
\end{equation}
The first sum is over collocation inputs $\{\boldsymbol{x}_q\}_{q=1}^{N_2}$ drawn across the full domain $\Omega$. The knowledge landmarks describe how the system behaves across the whole operating range, so the knowledge term is evaluated across that range. This lets the landmarks constrain the model outside $\Omega^\ast$, where no numeric observation is available. The points $\boldsymbol{x}_q$ carry no target value, so this step supplies no label information.

Uniform sampling would place most points where $A_i(\boldsymbol{x}_q) \approx 0$, so the gradients of the knowledge term vanish. We therefore sample the collocation points from Gaussians centered at the landmark prototypes $\boldsymbol{v}_i$. This keeps the knowledge term active while covering the operational domain $\Omega$. Across all experiments, we set $N_2 = 20000$ collocation points for model training. The term $V_i(\boldsymbol{x};\boldsymbol{a})$ is expressed as
\begin{equation}
V_i(\boldsymbol{x};\boldsymbol{a})
=
\operatorname{cov}\!\left(M(\boldsymbol{x};\boldsymbol{a}), B_i\right)\operatorname{sp}(B_i).
\label{eq:matching_term}
\end{equation}

For an active landmark with specificity $s_i = \operatorname{sp}(B_i) > 0$, the individual penalty $(1 - V_i)^2$ reaches its minimum $(1 - s_i)^2$ when the prediction coincides with the output context center, $M(\boldsymbol{x};\boldsymbol{a}) = c_i$. The specificity factor changes the curvature of the penalty without moving its minimum: a more specific context penalizes deviations from $c_i$ more steeply, whereas an uninformative context ($s_i = 0$) gives no gradient with respect to the prediction. Because $\operatorname{sp}(B_i) \le 1$, the penalty retains an offset $(1 - s_i)^2$, which grows as the context information granule becomes broader (less specific). Across the operational domain $\Omega$, several landmarks may be simultaneously active, and their activations $A_i(\boldsymbol{x}_q)$ combine these penalties into a soft regularizer that depends on the input and favors compatible output regions. In Eq.~\eqref{eq:augmented_loss}, the trade-off parameter $\lambda \in [0, 1]$ balances this global granular guidance against localized empirical fitting on $D^\ast$, motivating the validation protocol detailed below.

To make the combination of several active landmarks explicit, we treat the prediction $\hat{y}=M(\boldsymbol{x};\boldsymbol{a})$ at one input as a free variable. With $a_i=A_i(\boldsymbol{x})$ and $s_i=\operatorname{sp}(B_i)$, any stationary point of the point-wise penalty $\ell(\hat{y})=\sum_{i}a_i\bigl(1-s_iB_i(\hat{y})\bigr)^2$ satisfies
\begin{gather*}
\hat{y}^\ast=\frac{\sum_i \omega_i(\hat{y}^\ast)\,c_i}{\sum_i \omega_i(\hat{y}^\ast)},\\
\omega_i(\hat{y})=\frac{a_i s_i B_i(\hat{y})\bigl(1-s_iB_i(\hat{y})\bigr)}{\sigma_i^2}\ge 0.
\end{gather*}
The fixed point is a weighted average of the active context centers, so it lies between the smallest and the largest of them. When the active contexts overlap strongly, so that $B_i(\hat{y})\approx1$, Eq.~\eqref{eq:context_specificity} gives $\omega_i\approx\sqrt{\pi}\,a_is_i/[(y_{\mathrm{high}}-y_{\mathrm{low}})\,\sigma_i]$, so strongly activated and specific contexts receive the largest weights, while a context far from the prediction receives little weight through $B_i(\hat{y})$. For two equally activated contexts with centers $\bar{c}\pm d$, common width $\sigma$, and common specificity $s$, the midpoint $\bar{c}$ is a minimum if $r=d^2/\sigma^2\le1/2$ or $s>(2r-1)e^{r}/(4r-1)$; numerically, the critical separation $2d/\sigma$ lies between $1.4$ and $2.1$. Agreeing contexts are therefore averaged, whereas with strongly diverging contexts the prediction moves toward the context with the larger $a_i[1-(1-s_i)^2]$, which weights activation and specificity together.

\subsection{Validation Process}

Let $\boldsymbol{a}_{\lambda}$ denote the model parameters obtained by minimizing Eq.~\eqref{eq:augmented_loss} for a fixed trade-off parameter $\lambda$. The trained model is evaluated with two components. The first is the local data-fidelity error on $D^\ast$:
\begin{equation}
Q_1(\lambda)
=
\sqrt{
\frac{1}{N_1}
\sum_{k=1}^{N_1}
\left[
M(\boldsymbol{x}_k;\boldsymbol{a}_{\lambda})
-
\mathrm{target}_k
\right]^2 } .
\label{eq:q1}
\end{equation}
This term measures how well the model obtained under the fixed $\lambda$ preserves accuracy on the localized numeric observations. The second component is the full-domain validation error, computed on independently generated validation samples:
\begin{equation}
Q_{2,\mathrm{val}}(\lambda)
=
\sqrt{
\mathbb{E}_{\substack{\boldsymbol{x}\sim p^{\Omega}_{\mathrm{val}}(\boldsymbol{x})\\
\boldsymbol{w}\sim p_{\mathrm{val}}(\boldsymbol{w})}}
\left[
\left(M(\boldsymbol{x};\boldsymbol{a}_{\lambda})
-
f(\boldsymbol{x};\boldsymbol{w})\right)^2
\right]},
\label{eq:q2_val}
\end{equation}
where $p_{\mathrm{val}}(\boldsymbol{w})$ is defined over the admissible parameter intervals, and $p^{\Omega}_{\mathrm{val}}(\boldsymbol{x})$ is defined over the full input domain $\Omega$. The trade-off parameter is selected by minimizing
\begin{equation}
\widetilde{Q}_{\mathrm{val}}(\lambda)
=
Q_1(\lambda)+Q_{2,\mathrm{val}}(\lambda).
\label{eq:qval_total}
\end{equation}
For each candidate $\lambda$, we train the model with the augmented loss and compute Eq.~\eqref{eq:qval_total}. The selected $\lambda_{\mathrm{opt}}$ attains the minimum validation criterion.

Because $Q_{2,\mathrm{val}}(\lambda)$ uses simulator outputs over $\Omega$, this selection is an \emph{oracle reference procedure} for the main results in Tables~\ref{tab:env_results} and~\ref{tab:piston_results_kd}. In practical deployments where full-domain simulation is unavailable, $\lambda$ can be selected using label-free alternatives, such as monitoring the composite training loss or performing cross-validation with local observations and landmark compatibility. These practical selection rules and their performance are evaluated in Section~\ref{subsec:auto_lambda}.

After $\lambda_{\mathrm{opt}}$ has been determined, we evaluate the corresponding model on an independent full-domain test set:
\begin{equation}
Q_{2,\mathrm{test}}(\lambda_{\mathrm{opt}})
=
\sqrt{
\mathbb{E}_{\substack{\boldsymbol{x}\sim p^{\Omega}_{\mathrm{test}}(\boldsymbol{x})\\
\boldsymbol{w}\sim p_{\mathrm{test}}(\boldsymbol{w})}}
\left[
\left(M(\boldsymbol{x};\boldsymbol{a}_{\lambda_{\mathrm{opt}}})
-
f(\boldsymbol{x};\boldsymbol{w})\right)^2
\right]}.
\label{eq:q2_test}
\end{equation}
The full-domain validation and test samples are generated using different random seeds and contain no overlapping samples. Therefore, the full-domain test set is used neither for training the model parameters nor for selecting $\lambda$. %
In the experimental tables, we report the local fidelity $Q_1(\lambda_{\mathrm{opt}})$ and the full-domain accuracy $Q_{2,\mathrm{test}}(\lambda_{\mathrm{opt}})$ separately, because they describe different regions. We also report the normalized root mean squared error, $\mathrm{NRMSE} = Q_{2,\mathrm{test}} / (y_{\max} - y_{\min})$, with the output range computed over the full-domain test set.

The local observations $D^\ast$ are generated by one parameter vector $\boldsymbol{w}_0$, whereas Eqs.~\eqref{eq:q2_val} and~\eqref{eq:q2_test} draw a parameter vector for each sample. These equations define the \emph{mixture-of-worlds task}, whose target is the system family across scenarios; on it, even the simulator with $\boldsymbol{w}_0$ has a nonzero error. In the \emph{fixed-parameter task}, the test samples come from $\boldsymbol{w}_0$ itself, so the target is the full-domain response of the observed system. Section~\ref{sec:experiments} reports the mixture-of-worlds task unless stated otherwise; Section~\ref{subsec:auto_lambda} also reports the fixed-parameter task.

\subsection{Overall Procedure}
\label{subsec:procedure}

Algorithm~\ref{alg:kdml} summarizes the complete procedure. The landmarks $(c_i,\sigma_i,\operatorname{sp}(B_i),\boldsymbol{v}_{ij},\sigma_{ij}^{(d)})$ and the collocation inputs are built from $D$. The normalization factor $y_{\mathrm{span}}$ is computed from $D^\ast$. These quantities and the preset hyperparameters $C$, $K_i$, $m$, $\rho$, and $N_2$ are fixed before training. Only the network parameters $\boldsymbol{a}$ are optimized; $\lambda$ is chosen in Stage~3 among separately trained models. Inference requires only the query $\boldsymbol{x}$ and the trained parameters $\boldsymbol{a}_{\lambda_{\mathrm{opt}}}$.

\begin{algorithm}[!t]
\caption{Construction, Training, and Use of the \KD-Model}
\label{alg:kdml}
\begin{algorithmic}[1]
\REQUIRE Prior evidence $D$; local data $D^\ast$; $C$, $\{K_i\}$, $m$, $\rho$, $N_2$; candidate set $\Lambda\subset[0,1]$.
\ENSURE Predictor $M(\cdot\,;\boldsymbol{a}_{\lambda_{\mathrm{opt}}})$.
\STATE \textbf{Stage 1} (before training; $y_{\mathrm{span}}$ from $D^\ast$, all else from $D$):
\STATE Place $c_1,\ldots,c_C$ uniformly in $[y_{\mathrm{low}},y_{\mathrm{high}}]$; set each $\sigma_i$ by the $\kappa$th-nearest-neighbor rule and $\operatorname{sp}(B_i)$ by Eq.~\eqref{eq:context_specificity}.
\FOR{$i=1,\ldots,C$}
\STATE Run conditional FCM, Eqs.~\eqref{eq:cfcm_u}--\eqref{eq:cfcm_v}, to obtain $\boldsymbol{v}_{ij}$ and $u_{ijk}$, $j=1,\ldots,K_i$.
\STATE Select each width $\sigma_{ij}^{(d)}$ by Eq.~\eqref{eq:width_rule} and form $A_{ij}$ by Eq.~\eqref{eq:input_granule_tnorm}.
\ENDFOR
\STATE Flatten the pairs into $\mathsf{K}=\{(A_i,B_i)\}_{i=1}^{c}$.
\STATE Draw $N_2$ collocation inputs from Gaussians centered at the prototypes $\boldsymbol{v}_i$ with spreads $1.5\,\sigma_i^{(d)}$, clipped to $\Omega$.
\STATE \textbf{Stage 2} (training; $\mathsf{K}$ and collocation inputs fixed):
\FOR{each $\lambda\in\Lambda$}
\STATE Initialize $\boldsymbol{a}$; minimize $L(\boldsymbol{a};\lambda)$ in Eq.~\eqref{eq:augmented_loss} with Adam on mini-batches from $D^\ast$ and the collocation inputs, giving $\boldsymbol{a}_{\lambda}$.
\ENDFOR
\STATE \textbf{Stage 3} (selection of $\lambda$): in the benchmarks, minimize $\widetilde{Q}_{\mathrm{val}}(\lambda)$ in Eq.~\eqref{eq:qval_total}; in deployment, use a label-free rule such as CV-J (Section~\ref{subsec:auto_lambda}).
\STATE \textbf{Stage 4} (inference): for a query $\boldsymbol{x}\in\Omega$, return $M(\boldsymbol{x};\boldsymbol{a}_{\lambda_{\mathrm{opt}}})$.
\end{algorithmic}
\end{algorithm}

\section{Experimental Studies}
\label{sec:experiments}

We evaluate the proposed \KD-model on two physics-based benchmarks: Environmental Pollutant Dispersion~\cite{bliznyuk2008bayesian} and Piston Cycle-Time Simulation~\cite{kenett2021modern}. The predictive model $M(\boldsymbol{x};\boldsymbol{a})$ is a single-hidden-layer MLP with 64 $\tanh$ units, the size at which empirical performance saturated. Information granules are parameterized with Gaussian membership functions.

All models use the same training protocol with the Adam optimizer and mini-batches of 256 samples. To quantify run-to-run variability, each condition is evaluated with ten independent random seeds. Each seed sets both the dataset split and the network initialization. The trade-off parameter $\lambda$ is swept over a uniform grid of $50$ candidate values in $[0,1]$. Unless specified otherwise, we report the mean and standard deviation over the ten seeds and two-sided paired $t$-tests against the data-only baseline ($\lambda=1$).

\subsection{Environmental Pollutant Dispersion Benchmark}
\label{subsec:env_dispersion}

\begin{figure}[!t]
\centering
\subfloat[]{%
  \includegraphics[height=3.3cm]{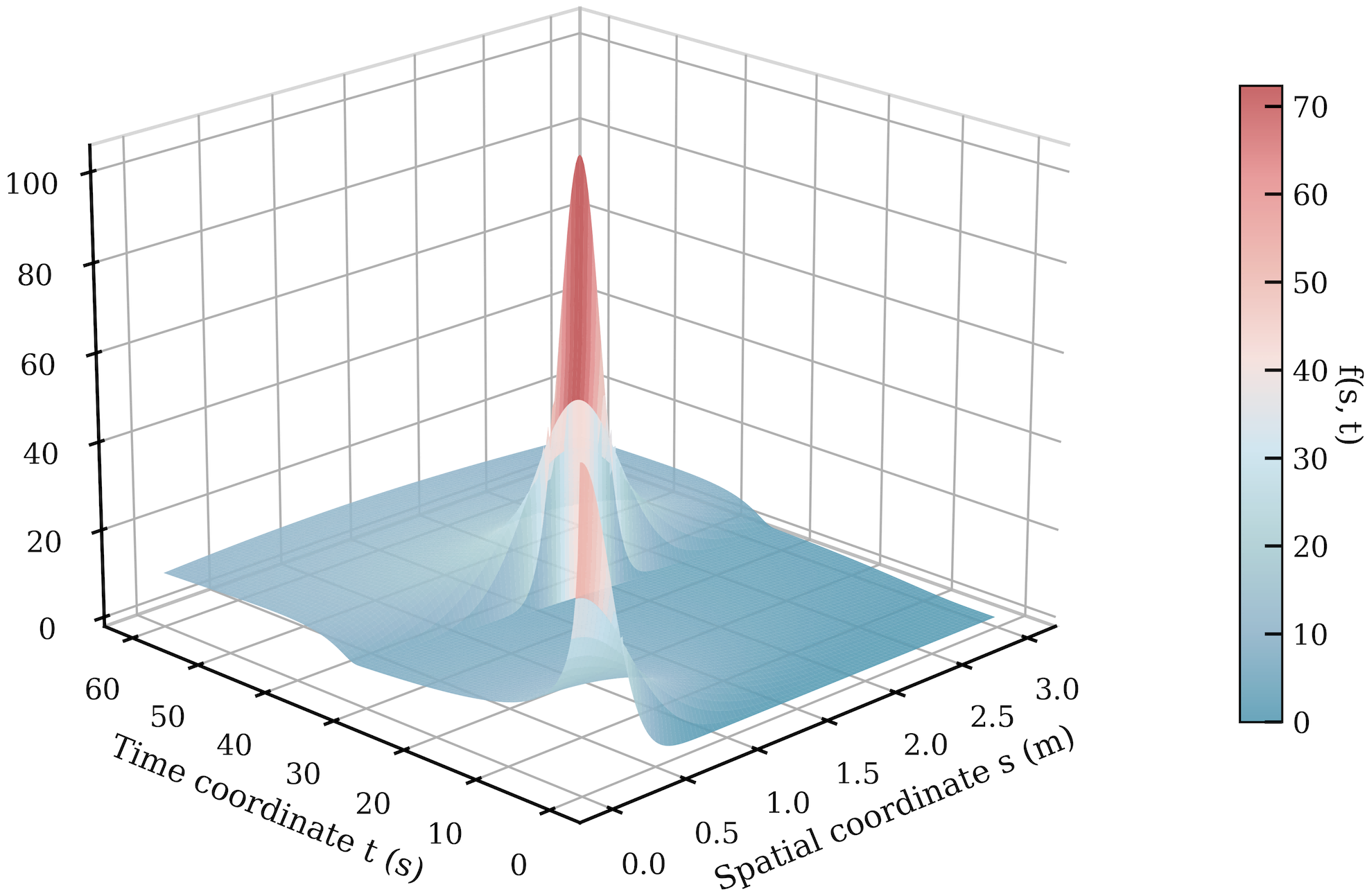}%
  \label{fig:env_domain_a}}
\hfill
\subfloat[]{%
  \includegraphics[height=3.3cm]{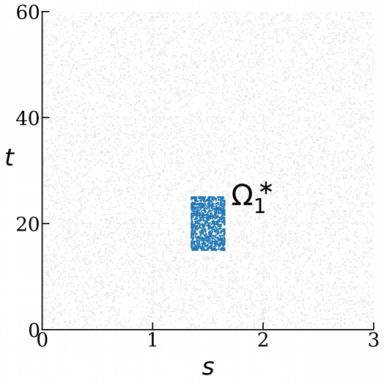}%
  \label{fig:env_domain_b}}
\caption{Environmental benchmark: (a) scaled concentration field $f(s,t)$ with baseline parameters and (b) domain $\Omega$ with local window $\Omega_1^\ast$.}
\label{fig:env_domain}
\end{figure}

An environmental dispersion model describes pollutant transport following a chemical spill. The analytical expression of the concentration field $C(s,t)$ at spatial coordinate $s\in[0,3]$ meters and time $t\in[0.3,60]$ seconds is given by
\begin{equation}
\begin{aligned}
C(s,t)
&=\frac{R}{\sqrt{4\pi Y t}}\exp\!\left(-\frac{s^2}{4Yt}\right) \\
&\quad +\frac{R}{\sqrt{4\pi Y (t-\tau)}}
\exp\!\left(-\frac{(s-L)^2}{4Y(t-\tau)}\right)\,\mathbb{I}(t>\tau),
\end{aligned}
\label{eq:env_c}
\end{equation}
where $\mathbb{I}(t>\tau)$ is the indicator function. The model output $f(s,t)=\sqrt{4\pi}\,C(s,t)$ is the scaled concentration over the spatiotemporal domain $\Omega=[0,3]\times[0.3,60]$. The parameter vector $\boldsymbol{W}=(R,Y,L,\tau)$ is sampled from uniform distributions over the ranges listed in Table~\ref{tab:env_parameters}. Fig.~\ref{fig:env_domain}(a) illustrates the scaled concentration field $f(s,t)$ across $\Omega$, evaluated with baseline parameters $R=10$ kg, $Y=0.07$ m$^2$/s, $L=1.505$ m, and $\tau=30.1525$ s. The full-domain splits are generated from $100$ parameter vectors drawn from the admissible intervals in Table~\ref{tab:env_parameters}, with $200$ points sampled per vector. The splits $D$, $D_{\mathrm{val}}$, and $D_{\mathrm{test}}$ each contain $20000$ labeled points. The knowledge landmarks are constructed from $D$ alone. The localized dataset $D^\ast$ contains $N_1=1000$ samples confined to the localized observation domain $\Omega^\ast \subset \Omega$.

\begin{table}[!t]
\caption{Variable Ranges and Parameters for Environmental Pollutant Dispersion}
\label{tab:env_parameters}
\centering
\small
\begin{tabular}{l c c c}
\toprule
Variable & Symbol & Range & Unit \\
\midrule
\multicolumn{4}{l}{\textit{Input Variables}} \\
Spatial Coordinate & $s$ & $[0,3]$ & m \\
Time & $t$ & $[0.3,60]$ & s \\
\midrule
\multicolumn{4}{l}{\textit{Parameters}} \\
Spilled Mass & $R$ & $[7,13]$ & kg \\
Diffusion Rate & $Y$ & $[0.02,0.12]$ & m$^2$/s \\
Second Spill Location & $L$ & $[0.01,3]$ & m \\
Second Spill Time & $\tau$ & $[30.01,30.295]$ & s \\
\bottomrule
\end{tabular}
\end{table}

\begin{figure}[!t]
\centering
\includegraphics[width=0.98\linewidth]{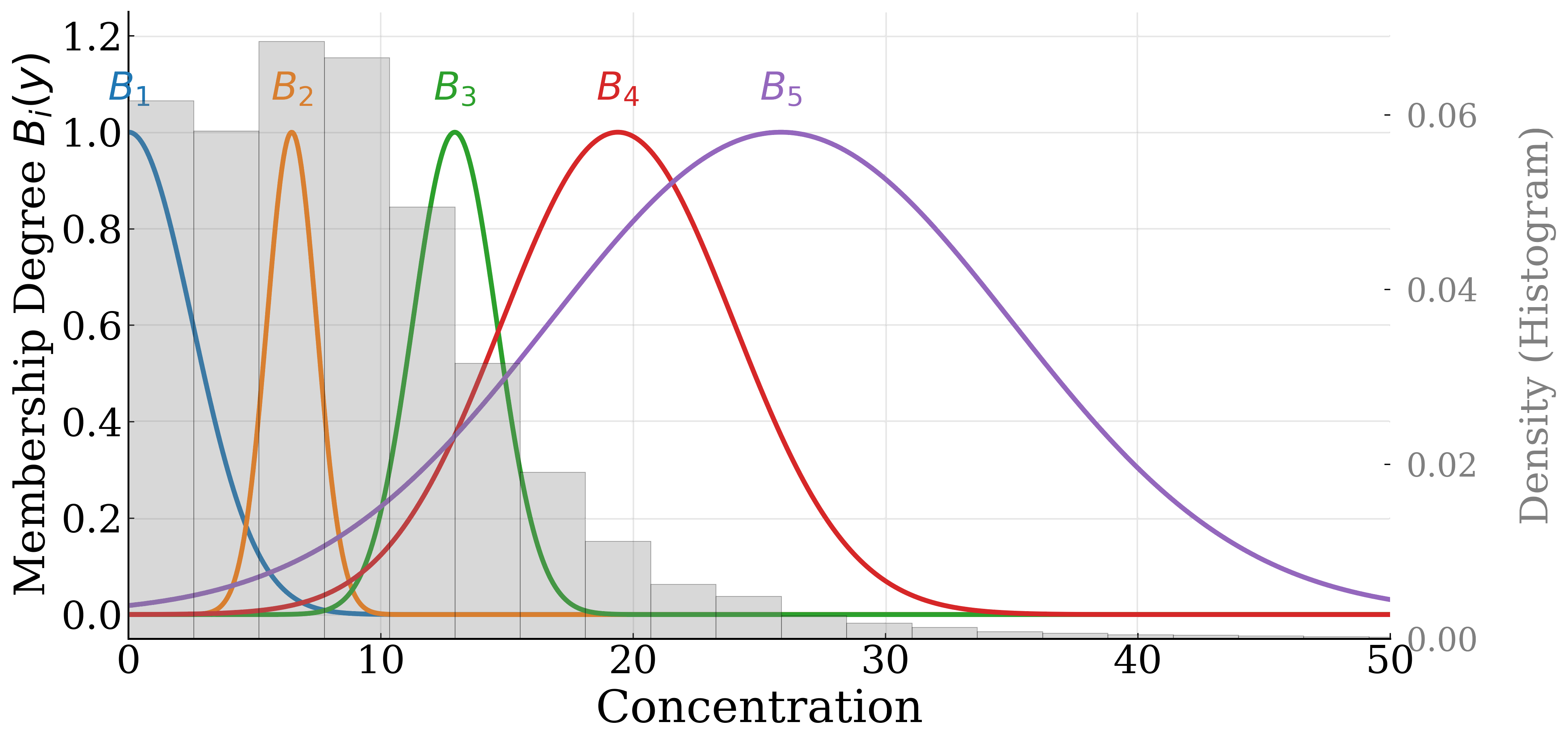}
\caption{Output contexts $B_i$ for the environmental benchmark, shown over the histogram of scaled concentrations in $\Omega$.}
\label{fig:env_output_granules}
\end{figure}

\begin{figure*}[!t]
\centering
\includegraphics[width=\textwidth]{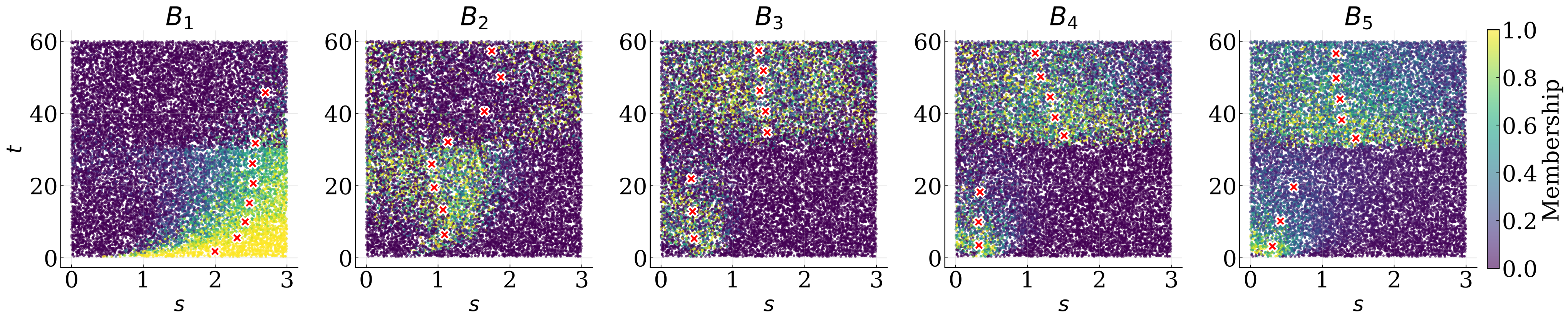}
\caption{Input granules $\{A_{ij}\}$ for the environmental model, obtained via conditional FCM for each output context. Each subplot shows eight input granules associated with one output context $B_i$, where the color of each point represents its membership degree.}
\label{fig:env_input_granules}
\end{figure*}

We construct $C=5$ output contexts and $K=8$ input clusters per context, producing $c=40$ knowledge landmarks $\mathsf{K}=\{(A_{ij},B_i)\mid i=1,\ldots,C,\ j=1,\ldots,K\}$ over $\Omega$. Output contexts $\{B_1,\ldots,B_5\}$ are formed using Gaussian membership functions (Fig.~\ref{fig:env_output_granules}) with context width parameter $\rho=0.2$. Conditional FCM then identifies $K=8$ input clusters in the $s$--$t$ space for each context. The resulting input granules are shown in Fig.~\ref{fig:env_input_granules}. Input granules for low-concentration contexts lie at early times or far from the source, whereas those for high-concentration contexts lie near the spill locations at later times, following the plume.

\begin{figure}[!t]
\centering
\subfloat[]{%
  \raisebox{1.4pt}{\includegraphics[width=0.45\linewidth,trim=15 3 13.5 9,clip]{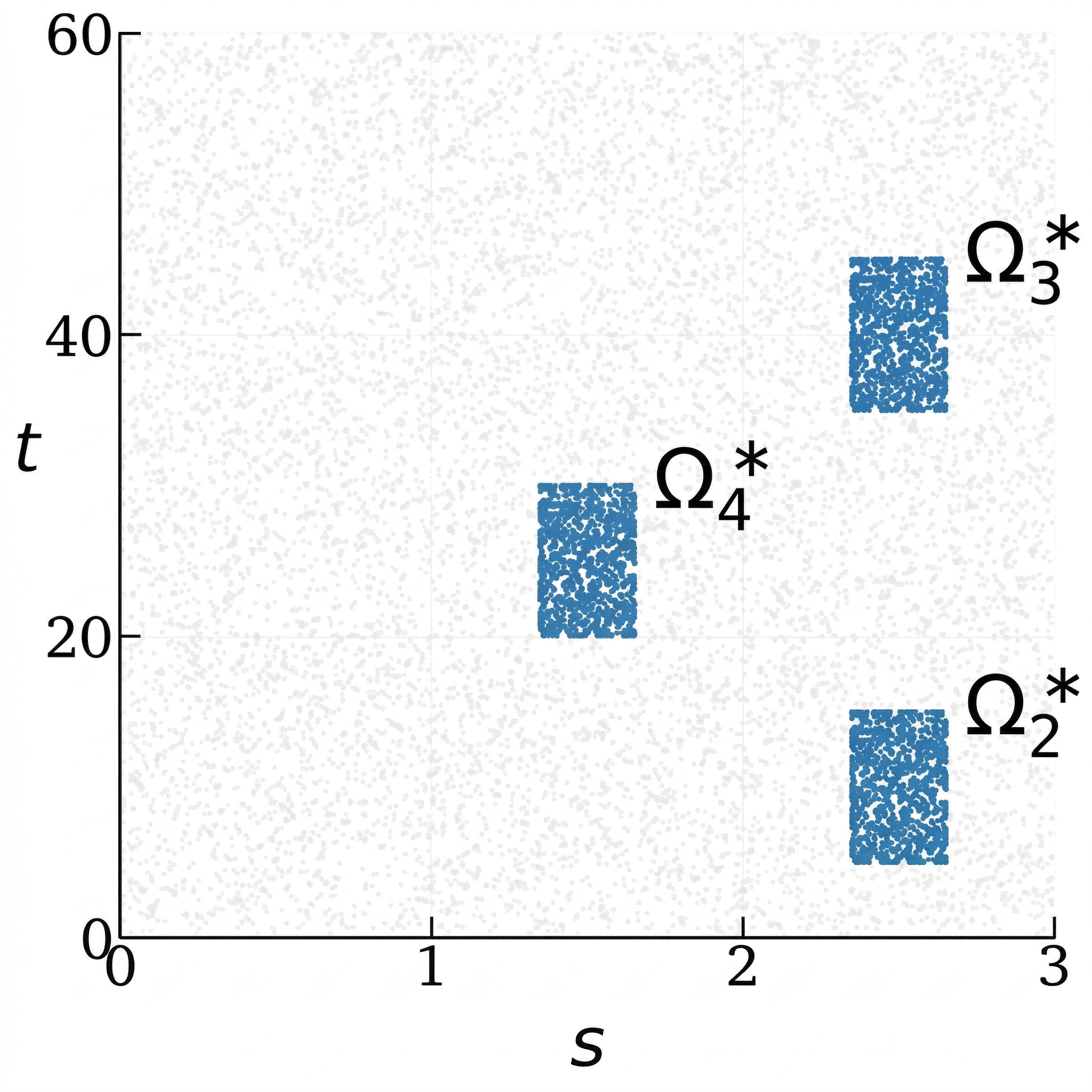}}%
  \label{fig:env_positions_a}}
\hfill
\subfloat[]{%
  \includegraphics[width=0.50\linewidth,trim=16.5 15 39 112.5,clip]{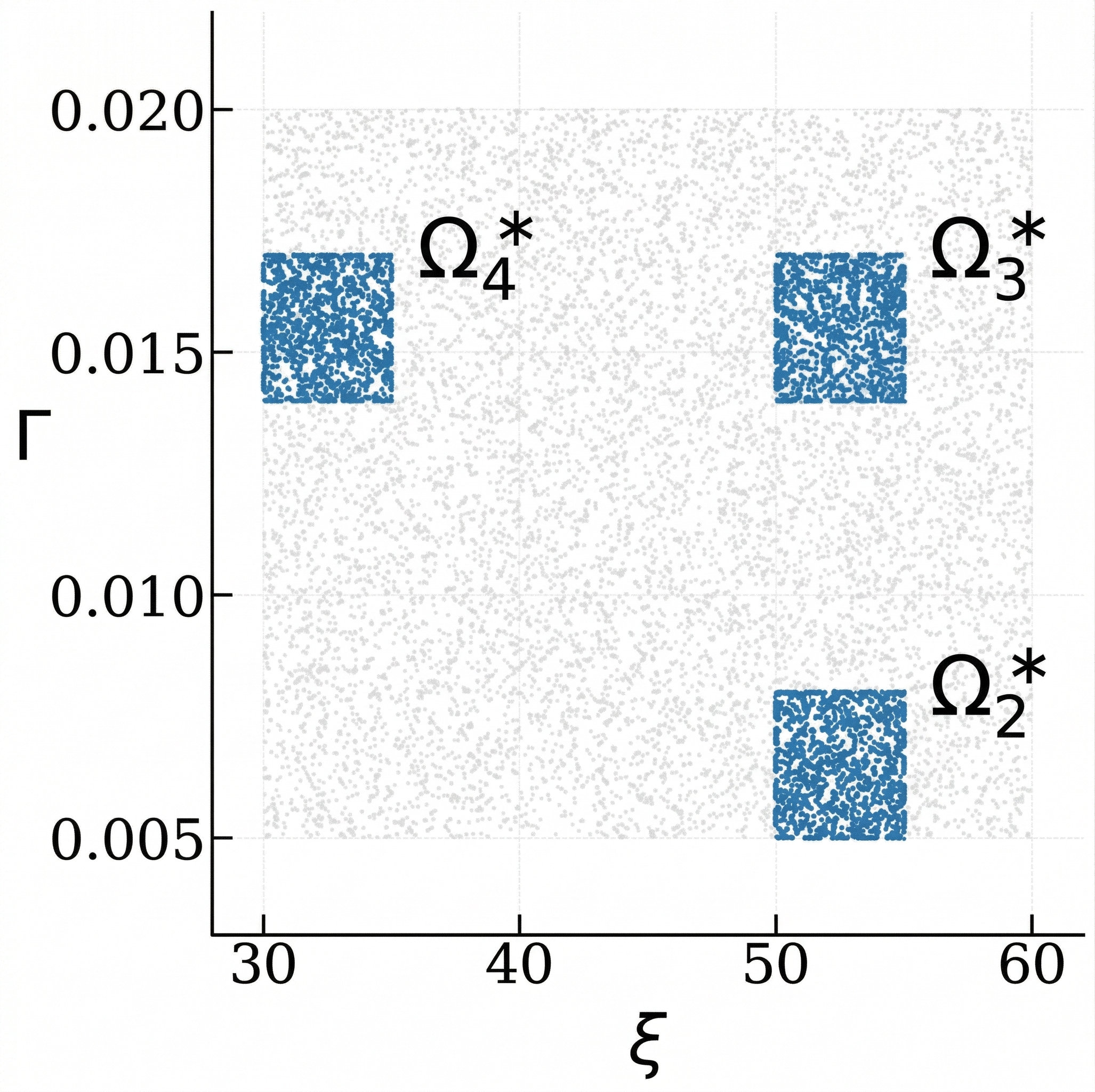}%
  \label{fig:piston_positions_b}}
\caption{Different positions of the local observation window $\Omega^{\ast}$ ($\Omega_2^\ast$, $\Omega_3^\ast$, and $\Omega_4^\ast$) for the two benchmarks. (a) Environmental benchmark. (b) Piston benchmark.}
\label{fig:positions}
\end{figure}

To evaluate model performance under diverse spatial observation constraints, we examine four localized observation windows $\Omega_1^\ast, \ldots, \Omega_4^\ast \subset \Omega$ (Fig.~\ref{fig:env_domain}(b) and Fig.~\ref{fig:positions}(a)):
\begin{itemize}
\item $\Omega_1^\ast$: default observation window ($s\in[1.35,1.65]$ m, $t\in[15,25]$ s), an intermediate transport stage;
\item $\Omega_2^\ast$($s\in[2.35,2.65]$ m, $t\in[5,15]$ s): early times and regions far from the spill source, where the concentration response is smooth and low in magnitude;
\item $\Omega_3^\ast$($s\in[2.35,2.65]$ m, $t\in[35,45]$ s): regions far from both spills in time and space, an uninformative downstream plateau;
\item $\Omega_4^\ast$($s\in[1.35,1.65]$ m, $t\in[20,30]$ s): just before the second spill activates, near its location.
\end{itemize}

Model training minimizes $L(\boldsymbol{a};\lambda)$ for $5000$ epochs at each candidate $\lambda$. The optimal parameter $\lambda_{\mathrm{opt}}$ is selected by minimizing $Q_1+Q_{2,\mathrm{val}}$ on the validation set, as described in Section~\ref{sec:lossfunction}. We evaluate generalization on the independent test set $D_{\mathrm{test}}$ with $Q_{2,\mathrm{test}}$; unless stated otherwise, this is the mixture-of-worlds task of Section~\ref{sec:lossfunction}. The absolute and relative reductions in full-domain test error over the data-only baseline ($\lambda=1$) are defined as
\begin{equation}
\Delta Q_2=Q_{2,\mathrm{test}}(1)-Q_{2,\mathrm{test}}(\lambda_{\mathrm{opt}}),
\label{eq:delta_q_abs}
\end{equation}
\begin{equation}
\Delta Q_2(\%)=\frac{\Delta Q_2}{Q_{2,\mathrm{test}}(1)}\times 100\%.
\label{eq:delta_q_rel}
\end{equation}
Unless stated otherwise, $\Delta Q_2(\%)$ is computed within each seed by Eq.~\eqref{eq:delta_q_rel} and averaged over the ten seeds; at environmental $\Omega_3^\ast$, this gives $24.89\%$, whereas the relative difference of the seed-averaged errors, $(8.285-6.086)/8.285$, is $26.5\%$.

\begin{figure}[!t]
\centering
\subfloat[]{%
  \includegraphics[width=0.49\linewidth]{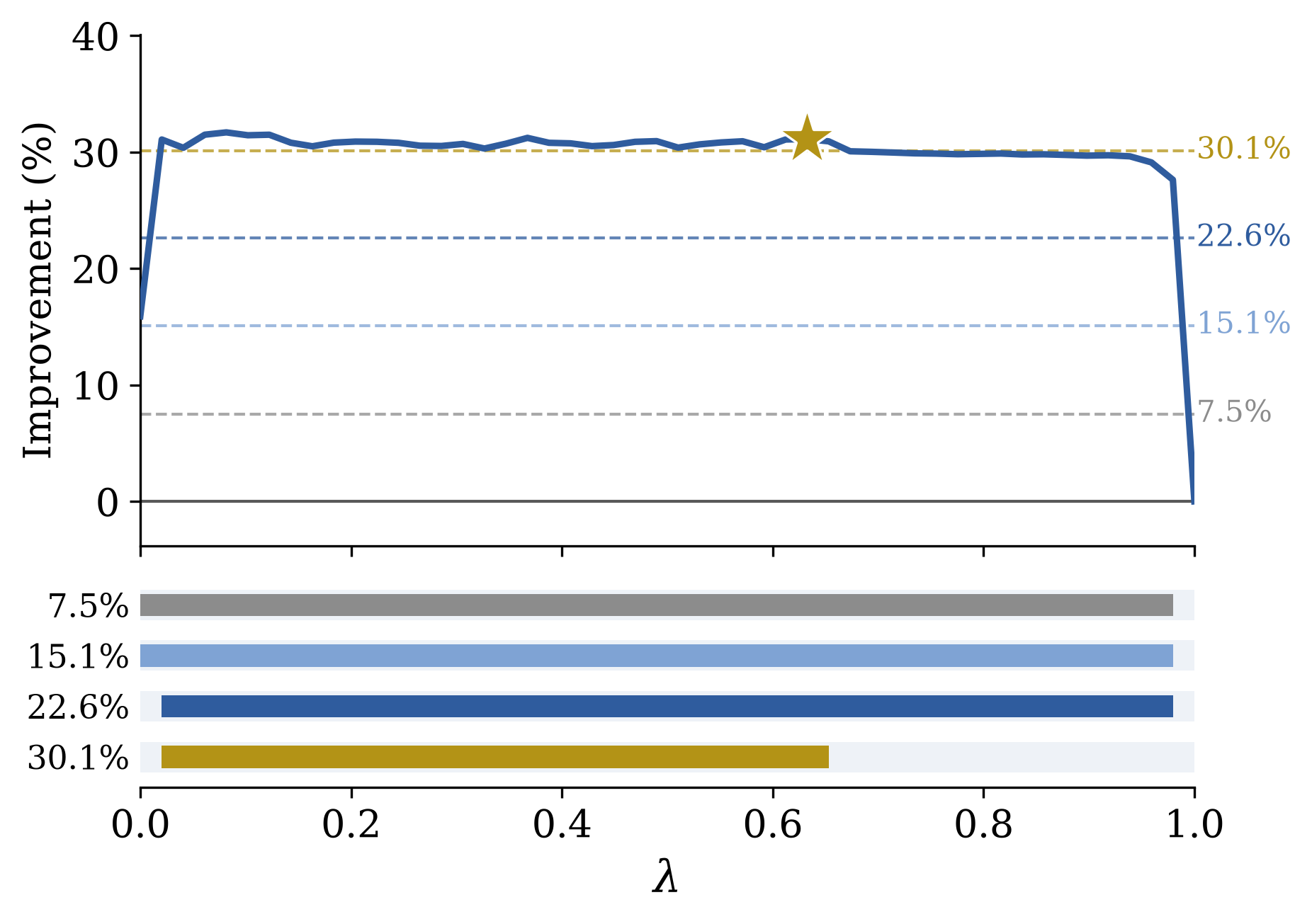}%
  \label{fig:lambda_gain_env_a}}
\hfill
\subfloat[]{%
  \includegraphics[width=0.49\linewidth]{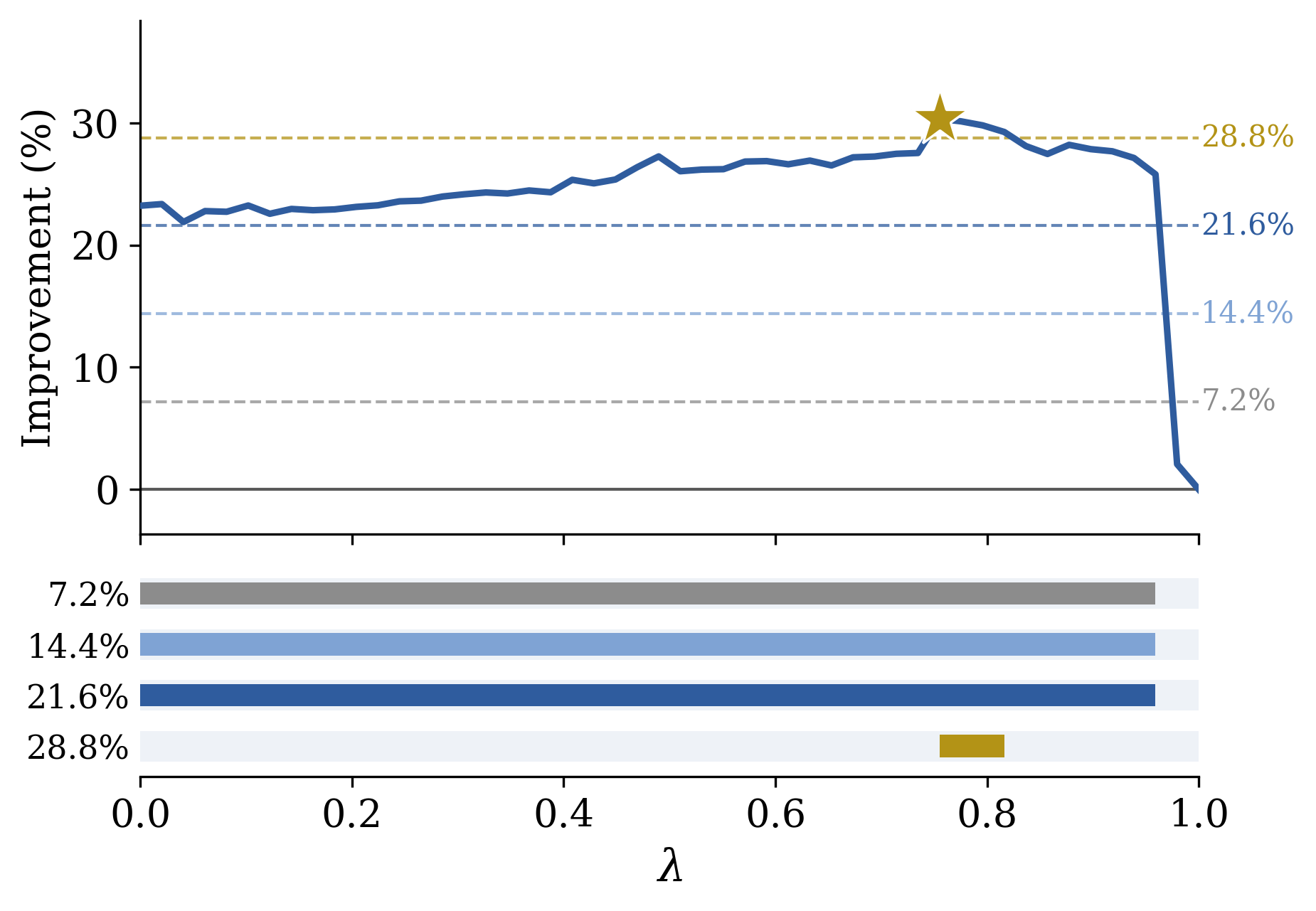}%
  \label{fig:lambda_gain_pis_b}}
\caption{Reduction in $Q_2$ over the data-only model as a function of $\lambda$ for one representative seed. The star marks the selected $\lambda$, and the lower bands show where the reduction exceeds each threshold. (a) Environmental benchmark, $\Omega_3^\ast$. (b) Piston benchmark, $\Omega_1^\ast$.}
\label{fig:lambda_gain_env}
\end{figure}

\begin{table*}[!t]
\centering

\caption{Results on the Environmental Benchmark ($^{\ast}$: $p<0.05$; Oracle $\lambda$)}
\label{tab:env_results}
\small
\begin{tabular*}{0.95\textwidth}{@{\extracolsep{\fill}} l l c c c c c @{}}
\toprule
$\Omega^\ast$ & Model & $Q_1$ & $Q_{2,\mathrm{val}}$ & $Q_{2,\mathrm{test}}$ & NRMSE & $\Delta Q_2(\%)$ \\
\midrule
\multirow{2}{*}{$\Omega_1^\ast$} & Data-only & 0.005 $\pm$ 0.002 & 7.661 $\pm$ 0.346 & 7.675 $\pm$ 0.379 & 0.051 & --- \\
 & \KD-model & 0.026 $\pm$ 0.026 & 7.255 $\pm$ 0.445 & 7.258 $\pm$ 0.554 & 0.048 & 5.47 $\pm$ 4.64$^{\ast}$ \\
\midrule
\multirow{2}{*}{$\Omega_2^\ast$} & Data-only & 0.006 $\pm$ 0.003 & 7.288 $\pm$ 0.395 & 7.300 $\pm$ 0.386 & 0.049 & --- \\
 & \KD-model & 0.015 $\pm$ 0.008 & 6.815 $\pm$ 0.517 & 6.834 $\pm$ 0.639 & 0.045 & 6.46 $\pm$ 5.95$^{\ast}$ \\
\midrule
\multirow{2}{*}{$\Omega_3^\ast$} & Data-only & 0.548 $\pm$ 0.315 & 8.319 $\pm$ 1.483 & 8.285 $\pm$ 1.464 & 0.055 & --- \\
 & \KD-model & 0.095 $\pm$ 0.038 & 6.063 $\pm$ 0.230 & 6.086 $\pm$ 0.231 & 0.041 & 24.89 $\pm$ 10.74$^{\ast}$ \\
\midrule
\multirow{2}{*}{$\Omega_4^\ast$} & Data-only & 0.006 $\pm$ 0.003 & 7.783 $\pm$ 0.348 & 7.797 $\pm$ 0.376 & 0.052 & --- \\
 & \KD-model & 0.032 $\pm$ 0.016 & 7.484 $\pm$ 0.328 & 7.505 $\pm$ 0.411 & 0.050 & 3.76 $\pm$ 1.41$^{\ast}$ \\
\bottomrule
\end{tabular*}
\end{table*}

Table~\ref{tab:env_results} reports full-domain accuracy for the four observation windows. The \KD-model reduces $Q_{2,\mathrm{test}}$ over the data-only baseline, with statistically significant improvements in each case. To interpret these error values, we evaluate three reference baselines on the same test set. A naive constant predictor that always outputs the test-set mean reaches $7.061$, the simulator with the exact parameters $\boldsymbol{w}_0$ of the local observations reaches $6.131$, and the theoretical conditional expectation given $\boldsymbol{x}$ alone, which is the optimal Bayes lower bound for this task, reaches $4.945$. The data-only baseline yields errors between $7.300$ and $8.285$, all worse than the constant predictor; without guidance, it does not extrapolate reliably outside $\Omega^\ast$.

The gain is largest at $\Omega_3^\ast$, where the \KD-model reduces $Q_{2,\mathrm{test}}$ by $24.89\%$ per seed and closes the gap to the physical reference model. Because $\Omega_3^\ast$ lies on a flat downstream plateau, local observations contain little information about the source dynamics.  Across $\Omega_1^\ast, \Omega_2^\ast$, and $\Omega_4^\ast$, the \KD-model reduces $Q_{2,\mathrm{test}}$ by $5.47\%, 6.46\%$, and $3.76\%$, closing $18\%$--$40\%$ of the gap to the physical reference. The gain is smallest at $\Omega_4^\ast$, which lies just before the second spill activates.

Fig.~\ref{fig:lambda_gain_env}(a) displays the sensitivity of error reduction to $\lambda$ at $\Omega_3^\ast$ for a representative run. Across the ten seeds, error reductions remain positive over more than $80\%$ of the candidate grid for $\Omega_1^\ast$, $\Omega_3^\ast$, and $\Omega_4^\ast$, whereas $\Omega_2^\ast$ yields positive reductions over $26.8\%$ of the grid. At $\Omega_3^\ast$, reductions exceeding $10\%$ occur across $80.0\%$ of the grid. The gain therefore holds over broad ranges of $\lambda$ at most observation windows, while $\Omega_2^\ast$ benefits from a finer search.

\subsection{Piston Cycle-Time Simulation Benchmark}
\label{subsec:piston_benchmark}

\begin{figure}[!t]
\centering
\subfloat[]{%
  \includegraphics[height=3.5cm]{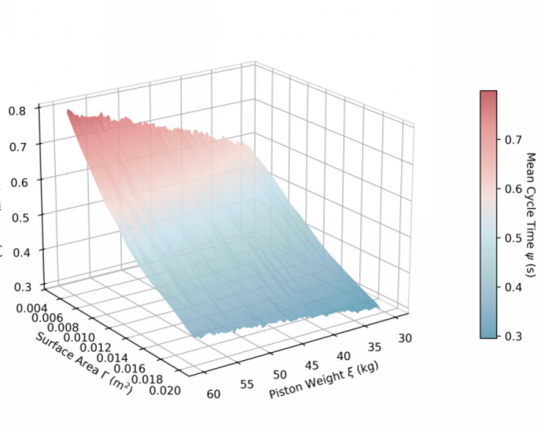}%
  \label{fig:piston_domain_a}}
\hfill
\subfloat[]{%
  \includegraphics[height=3.5cm]{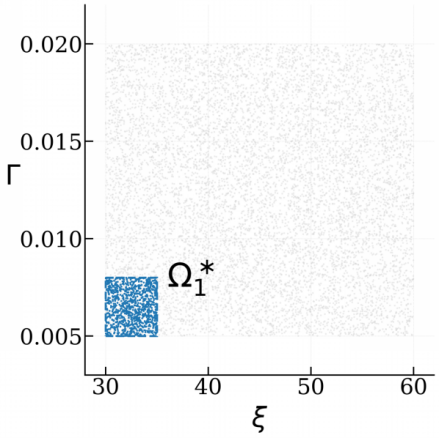}%
  \label{fig:piston_domain_b}}
\caption{Piston benchmark design space: (a) cycle-time surface $\psi(\xi,\Gamma)$ and (b) domain $\Omega$ with local window $\Omega_1^\ast$.}
\label{fig:piston_domain_kd}
\end{figure}

\predisplaypenalty=0
The piston simulation models cycle time $\psi$ as a function of piston weight $\xi$ and surface area $\Gamma$:
\begin{equation}
\psi(\xi,\Gamma)=2\pi\sqrt{\frac{\xi}{k+\frac{\Gamma^2 P_0 V_0 T_a}{T_0 V^2}}},
\label{eq:piston_psi}
\end{equation}\predisplaypenalty=10000
with intermediate gas volume $V$ and force term $A$ specified by
\begin{equation}
\begin{aligned}
V &= \frac{\Gamma}{2k}\left(\sqrt{A^2+\frac{4kP_0V_0T_a}{T_0}}-A\right), \\
A &= P_0\Gamma+19.62\,\xi-\frac{kV_0}{\Gamma},
\end{aligned}
\label{eq:piston_aux}
\end{equation}
over the domain $\Omega=[30,60]\times[0.005,0.020]$. The parameter vector $\boldsymbol{W}=(V_0,k,P_0,T_a,T_0)$ is drawn uniformly from the intervals in Table~\ref{tab:piston_parameters_kd}. Fig.~\ref{fig:piston_domain_kd}(a) shows the cycle-time response surface, and Fig.~\ref{fig:piston_domain_kd}(b) illustrates the local observation region $\Omega_1^\ast$ within $\Omega$.

\begin{table}[!t]
\caption{Variable Ranges and Parameters for Piston Simulation}
\label{tab:piston_parameters_kd}
\centering
\small
\begin{tabular}{l c c c}
\toprule
Variable & Symbol & Range & Unit \\
\midrule
\multicolumn{4}{l}{\textit{Input Design Variables}} \\
Piston Weight & $\xi$ & $[30,60]$ & kg \\
Surface Area & $\Gamma$ & $[0.005,0.020]$ & m$^2$ \\
\midrule
\multicolumn{4}{l}{\textit{Parameters}} \\
Gas Volume & $V_0$ & $[0.002,0.010]$ & m$^3$ \\
Spring Coeff. & $k$ & $[1000,5000]$ & N/m \\
Atm. Pressure & $P_0$ & $[9\times 10^4,1.1\times 10^5]$ & N/m$^2$ \\
Ambient Temp. & $T_a$ & $[290,296]$ & K \\
Filling Temp. & $T_0$ & $[340,360]$ & K \\
\bottomrule
\end{tabular}
\end{table}

The knowledge base consists of $c=25$ landmarks constructed from $C=5$ output contexts and $K=5$ input clusters per context. The corresponding clustering results, including the output contexts and input membership heatmaps, are visualized in Appendix~A, Figs.~10--11. The prototypes follow the physical trend: cycle time increases with piston weight. Model training follows the same protocol, minimizing $L(\boldsymbol{a};\lambda)$ for $4000$ epochs across $50$ candidate values of $\lambda$.

We examine four operating windows $\Omega_1^\ast, \ldots, \Omega_4^\ast \subset \Omega$ (Fig.~\ref{fig:piston_domain_kd}(b) and Fig.~\ref{fig:positions}(b)):
\begin{itemize}
\item $\Omega_1^\ast$: default localized design region ($\xi\in[30,35]$ kg, $\Gamma\in[0.005,0.008]$ m$^2$);
\item $\Omega_2^\ast$: heavy piston with small surface area ($\xi\in[50,55]$ kg, $\Gamma\in[0.005,0.008]$ m$^2$), where high inertia lengthens the cycle and reduces sensitivity to $\Gamma$;
\item $\Omega_3^\ast$: heavy piston with large surface area ($\xi\in[50,55]$ kg, $\Gamma\in[0.014,0.017]$ m$^2$), where inertial and pressure forces are both large;
\item $\Omega_4^\ast$: light piston with large surface area ($\xi\in[30,35]$ kg, $\Gamma\in[0.014,0.017]$ m$^2$), where the response depends strongly and nonlinearly on $\Gamma$.
\end{itemize}

\begin{table*}[!t]
\centering

\caption{Results on the Piston Benchmark ($^{\ast}$: $p<0.05$; Oracle $\lambda$)}
\label{tab:piston_results_kd}
\small
\begin{tabular*}{0.95\textwidth}{@{\extracolsep{\fill}} l l c c c c c @{}}
\toprule
$\Omega^\ast$ & Model & $Q_1$ & $Q_{2,\mathrm{val}}$ & $Q_{2,\mathrm{test}}$ & NRMSE & $\Delta Q_2(\%)$ \\
\midrule
\multirow{2}{*}{$\Omega_1^\ast$} & Data-only & 0.0008 $\pm$ 0.0003 & 0.1904 $\pm$ 0.0661 & 0.1938 $\pm$ 0.0692 & 0.204 & --- \\
 & \KD-model & 0.0004 $\pm$ 0.0001 & 0.1160 $\pm$ 0.0073 & 0.1106 $\pm$ 0.0078 & 0.116 & 37.92 $\pm$ 16.14$^{\ast}$ \\
\midrule
\multirow{2}{*}{$\Omega_2^\ast$} & Data-only & 0.0008 $\pm$ 0.0004 & 0.1991 $\pm$ 0.0606 & 0.1979 $\pm$ 0.0648 & 0.208 & --- \\
 & \KD-model & 0.0005 $\pm$ 0.0003 & 0.1174 $\pm$ 0.0052 & 0.1137 $\pm$ 0.0059 & 0.119 & 38.29 $\pm$ 14.65$^{\ast}$ \\
\midrule
\multirow{2}{*}{$\Omega_3^\ast$} & Data-only & 0.0005 $\pm$ 0.0002 & 0.1633 $\pm$ 0.0302 & 0.1669 $\pm$ 0.0320 & 0.175 & --- \\
 & \KD-model & 0.0008 $\pm$ 0.0005 & 0.1163 $\pm$ 0.0188 & 0.1117 $\pm$ 0.0225 & 0.117 & 31.67 $\pm$ 13.61$^{\ast}$ \\
\midrule
\multirow{2}{*}{$\Omega_4^\ast$} & Data-only & 0.0004 $\pm$ 0.0002 & 0.2392 $\pm$ 0.0724 & 0.2470 $\pm$ 0.0749 & 0.258 & --- \\
 & \KD-model & 0.0006 $\pm$ 0.0001 & 0.1254 $\pm$ 0.0233 & 0.1212 $\pm$ 0.0293 & 0.127 & 47.95 $\pm$ 16.56$^{\ast}$ \\
\bottomrule
\end{tabular*}
\end{table*}

Table~\ref{tab:piston_results_kd} reports full-domain accuracy across all windows. The \KD-model consistently reduces full-domain test error $Q_{2,\mathrm{test}}$ on every seed and window, with statistically significant average reductions of $37.92\%$, $38.29\%$, $31.67\%$, and $47.95\%$. On the same mixture-of-worlds test set, the simulator with the exact parameters $\boldsymbol{w}_0$ has $Q_{2,\mathrm{test}}=0.0868$, and the \KD-model closes $69\%$ to $79\%$ of the gap between the data-only model and the simulator. The \KD-model is also more stable across seeds: the standard deviation of $Q_{2,\mathrm{test}}$ is $0.006$--$0.029$, against $0.032$--$0.075$ for the data-only baseline. Fig.~\ref{fig:lambda_gain_env}(b) shows a representative run at $\Omega_1^\ast$. The reduction is above $21.6\%$ for $\lambda\in[0,0.96]$. It reaches $30.3\%$ at the selected $\lambda\approx0.76$ and falls near $\lambda=1$.

\subsection{Diagnostic Controls and Baseline Validity}
\label{subsec:controls_baseline}

The improvement of the \KD-model over the data-only model could come from the network, from the full-domain inputs used to build the landmarks, or from the landmarks themselves. To test the first, we fit a degree-three ridge polynomial on the same local data $D^\ast$. As Table~\ref{tab:nonneural} shows, its full-domain error reaches $63.019$ at environmental $\Omega_2^\ast$ and $58.802$ at $\Omega_3^\ast$, so the extrapolation failure at these windows is not specific to the network.

\begin{table}[!t]
\centering

\caption{Full-Domain Error $Q_{2,\mathrm{test}}$ on the Environmental Benchmark for Models Fitted on $D^\ast$ Alone, Averaged Over Ten Seeds}
\label{tab:nonneural}
\small
\begin{tabular}{l c c c c}
\toprule
Model & $\Omega_1^\ast$ & $\Omega_2^\ast$ & $\Omega_3^\ast$ & $\Omega_4^\ast$ \\
\midrule
Data-only network & 7.675 & 7.300 & 8.285 & 7.797 \\
Polynomial, degree 3 & 7.276 & 63.019 & 58.802 & 7.609 \\
\bottomrule
\end{tabular}
\end{table}

On environmental $\Omega_3^\ast$, two controls separate the other two sources. They share the network, optimizer, and training budget of the \KD-model, and both see the $N_D=20000$ full-domain inputs used to build the landmarks. The self-training control retrains for three rounds on its own predictions and uses no output knowledge. It lowers the seed-averaged error of the data-only model by only $8.2\%$, and its error stays above that of a constant predictor at the test mean. The \KD-model lowers that error by $26.5\%$, so full-domain input coverage alone does not reproduce the gain.

The coarse-supervision control uses the input granules $A_{ij}$ and the context centers $c_i$, without the context widths or the output specificity, and collapses them into one target per input. At this window its test error is $6.362$, against $6.086$ for the \KD-model. This difference is significant in the paired tests of Appendix~B, so the full landmark term is more accurate than a single coarse target built from the same granules. The comparison is limited to environmental $\Omega_3^\ast$.

Within the landmark term, we then separate the PJG input widths from the output specificity at environmental $\Omega_3^\ast$ and piston $\Omega_1^\ast$, with ten seeds per window. Each design is removed on its own and both are removed together. With both designs removed, the simplified landmark term retains $93.0\%$ and $84.3\%$ of the reduction over the data-only model, so most of the gain comes from the landmarks themselves. The simplified term has $2.5\%$ and $10.8\%$ higher error than the full model. The full model is better on nine of the ten environmental seeds and on all ten piston seeds. The specificity contributes more at environmental $\Omega_3^\ast$, and the PJG widths more at piston $\Omega_1^\ast$. Appendix~F reports the full comparison.

\subsection{Robustness to Imperfect and Perturbed Knowledge}
\label{subsec:landmark_robustness}

Domain knowledge is rarely exact: prior estimates may be inaccurate, have imprecise spatial boundaries, or omit rules. To examine whether the \KD-model tolerates imperfect knowledge rather than requiring exact knowledge, we apply random corruptions to the landmark parameters. At windows $\Omega_1^\ast$ and $\Omega_3^\ast$ of each benchmark, the local observations $D^\ast$ and the test sets are held fixed over ten runs, while supplied landmarks are subjected to three types of corruption: (i)~\emph{random output context center shifts} with $\delta \in \{0.05, 0.10, 0.20\}$, zero-mean and independent across contexts; (ii)~\emph{random input prototype displacements} with $\delta \in \{0.05, 0.10, 0.20\}$, reflecting imprecise domain localization and boundary uncertainty; and (iii)~\emph{random landmark deletion} with probability $p \in \{0.10, 0.25, 0.50\}$ per landmark, which simulates an incomplete knowledge base.

With corrupted landmarks, the \KD-model outperforms the data-only baseline in $35$ of the $36$ evaluated conditions. Under input prototype displacements at $\delta=0.20$, the average degradation relative to the unperturbed model is $4.91\%$. Under landmark deletion at $p=0.50$, the model remains better than data-only across all four benchmark configurations, with an average degradation of $13.68\%$. Output-center shifts have the strongest effect, leading to an average degradation of $19.36\%$ at $\delta=0.20$, yet $11$ of the $12$ conditions still improve upon the baseline. One explanation is that the spatial overlap of the information granules allows neighboring landmarks to maintain regional constraints when individual granules are corrupted or missing. Appendix~C gives detailed formulations and results for each condition.

\subsection{Trade-Off Selection from Local Observations and Landmarks}
\label{subsec:auto_lambda}

In practice, the trade-off parameter $\lambda$ must be selected without oracle full-domain validation labels. Standard cross-validation on $D^\ast$ is unsuitable because it evaluates error only within $\Omega^\ast$ and therefore selects $\lambda$ near $1$, so the landmarks have little effect. To resolve this, we introduce two-fold cross-validation with a joint score (CV-J). For each fold $f \in \{1,2\}$, let $T_f$ and $V_f$ denote the training and validation splits of $D^\ast$. A candidate model $M_{\lambda,-f}$ is trained on $T_f$ alongside the landmarks, and the trade-off is chosen to minimize
\begin{equation}
\begin{aligned}
S_{\mathrm{CV}}(\lambda) &= \frac{1}{2}\sum_{f=1}^{2} \left[ \frac{\sum_{(\boldsymbol{x},y)\in V_f} \bigl(M_{\lambda,-f}(\boldsymbol{x})-y\bigr)^2}{|V_f|r_f^2} \right. \\
&\hspace{12mm}\left. +\; L_{\mathrm{knowledge},U_f^{\rm ref}}(M_{\lambda,-f}) \right],
\end{aligned}
\label{eq:cvj_score}
\end{equation}
where $r_f$ is the local target range in $T_f$, and $U_f^{\rm ref}$ contains $4096$ unlabeled collocation points sampled across $\Omega$. The final model is then refitted on all of $D^\ast$ using the selected trade-off.

We evaluate CV-J at environmental $\Omega_3^\ast$ and piston $\Omega_1^\ast$ under clean, sparse, and noisy local data, in runs separate from those of Tables~\ref{tab:env_results} and~\ref{tab:piston_results_kd}. Across all $60$ evaluated cases spanning the six conditions and ten repetitions, CV-J achieves lower error than the data-only baseline. On the fixed-parameter task, its mean per-seed error reductions are $29.7\%\text{--}32.7\%$ on environmental and $52.7\%\text{--}60.2\%$ on piston. Standard local cross-validation selects $\lambda=1$ on the piston data because it evaluates only within $\Omega^\ast$, yielding an RMSE of $0.194$. In contrast, CV-J selects $\lambda<1$ and reduces the error to $0.090$. A fixed default of $\lambda=0.5$ performs comparably to CV-J in these six conditions. Across all eight windows of Tables~\ref{tab:env_results} and~\ref{tab:piston_results_kd}, the simpler label-free rule Training~J, which minimizes the sum of the normalized data and knowledge losses at the end of training, retains $90.8\%$--$96.6\%$ of the improvement obtained with the oracle reference selection on the environmental benchmark and $82.7\%$--$98.0\%$ on the piston benchmark. Detailed score formulations and complete multi-condition comparisons are reported in Appendix~D.

\subsection{Sensitivity to Observation Noise}
\label{subsec:noisy_supervision}

Local observations can be corrupted by sensor degradation and measurement fluctuations. To assess tolerance to label noise, we inject zero-mean Gaussian noise into local training targets on the piston benchmark at $\Omega_1^\ast$. The corrupted targets $\widetilde{y}_k$ are generated as
\begin{equation}
\widetilde{y}_k = y_k + \epsilon_k, \quad \epsilon_k \sim \mathcal{N}(0, \sigma^2), \quad \sigma = \alpha \,\mathrm{std}\left(\{y_k\}_{k=1}^{N_1}\right),
\label{eq:noise_model}
\end{equation}
where $\alpha \in \{0, 0.1, \dots, 0.9\}$ controls the noise intensity. Each noise level is run ten times with $N_1=1000$, and the test targets and landmarks remain clean. In this study, $\lambda$ is selected for each run by the label-free training-loss rule (Training~J) over $\{0,0.1,\ldots,1\}$.

\begin{figure}[!t]
\centering
\includegraphics[width=0.98\linewidth]{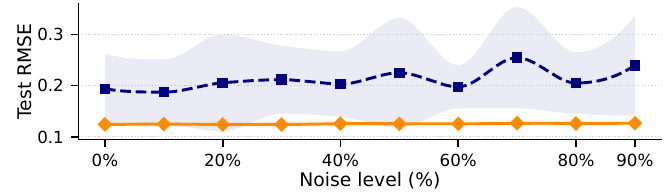}
\caption{Test RMSE under noisy local supervision on the piston benchmark at $\Omega_1^\ast$. Markers denote means over ten runs, and shaded bands show one sample standard deviation. Curves interpolate between evaluated noise levels. The dashed line with squares denotes the data-only model, and the solid line with diamonds denotes the \KD-model.}
\label{fig:noise_error}
\end{figure}

As shown in Fig.~\ref{fig:noise_error}, the data-only baseline is sensitive to label noise: its test error rises from $0.1938$ to $0.2396$, with large variance across runs. The \KD-model changes little: its test error shifts by less than $0.003$, from $0.1242$ at $\alpha=0$ to $0.1267$ at $\alpha=0.9$, and varies little across runs. At $\alpha=0.9$, where the noise variance is $81\%$ of the clean-target variance, the mean test RMSE of the \KD-model is only $2.0\%$ above its noise-free value. The landmarks therefore regularize the model and limit overfitting to the corrupted local targets.

\subsection{Sensitivity to Granule Resolution}
\label{subsec:granule_resolution}
\label{subsec:landmark_resolution}

We further examine sensitivity to the granularity of the landmark representation, governed by the number of output contexts $C$ and input granules per context $K$. We evaluate environmental $\Omega_1^\ast$ across $C \in \{3, 5, 7\}$ and $K \in \{4, 8, 12\}$ over three independent runs per configuration. At fixed $C$, varying the input partition $K$ changes mean test RMSE by less than $0.1$, suggesting that performance is not sensitive to the number of input clusters in this window. The number of output contexts $C$ has a greater influence. In all nine evaluated configurations, the mean test error is lower than that of the data-only baseline, with reductions between $4.4\%$ and $17.8\%$. These results suggest that, at environmental $\Omega_1^\ast$, the benefit does not depend on a finely tuned granulation. Since both benchmarks have only two inputs, Appendix~E examines how landmark activations change as the input dimension increases.

\section{Conclusion}
\label{sec:conclusion}

In this study, we introduced the \KD\ framework for learning from localized numeric data and coarse, non-equational domain knowledge. The framework represents this knowledge as granular landmarks, which enter the training objective as differentiable compatibility terms and complement local data fitting. In the two benchmarks, the landmarks improved predictions outside $\Omega^\ast$ without requiring governing equations or boundary conditions. The method was also robust to imperfect knowledge: it remained more accurate than the data-only baseline in $35$ of the $36$ tested conditions with randomly deleted or shifted landmarks. The trade-off between data and knowledge can be selected without full-domain labels while retaining most of the gain.

There are several directions worth exploring in future work. The trade-off could be adapted spatially using local model uncertainty, so that the model relies more on prior knowledge where data are scarce. The formulation could be extended to combine domain sources that differ in coverage and precision. A third direction is to integrate the granular structure into the network architecture, so that knowledge and representation are learned together.

\end{document}